\documentclass[final,3p,times]{elsarticle}

\usepackage{amssymb}
\usepackage{amsmath}
\usepackage{natbib}
\usepackage{orcidlink}
\usepackage[all]{hypcap}
\usepackage{hyperref} 
\usepackage{amsfonts}
\usepackage{array}
\usepackage[caption=false,font=normalsize,labelfont=sf,textfont=sf]{subfig}
\usepackage{textcomp}
\usepackage{stfloats}
\usepackage{url}
\usepackage{float}      
\usepackage{booktabs}  
\usepackage{tabularx}
\usepackage{tikz,xcolor}      
\usepackage{orcidlink}
\usepackage{threeparttable}
\usepackage{multirow}
\usepackage{lscape}
\usepackage{pifont}
\usepackage{algorithm}
\usepackage{algorithmic}
\usepackage{stfloats}

\hypersetup{hidelinks,
colorlinks=true,
allcolors=black,
pdfstartview=Fit,
breaklinks=true}

\journal{Neurocomputing}
\begin{document}

\begin{frontmatter}
\title{Bridging the Gaps: Utilizing Unlabeled Face Recognition Datasets to Boost Semi-Supervised Facial Expression Recognition}

\author[add1]{Jie Song\orcidlink{0009-0007-1606-0605}}
\ead{songjie02\_09@163.com}

\author[add1]{Mengqiao He\orcidlink{0000-0003-1900-4178}}
\ead{hmq0930@163.com}

\author[add1]{Jinhua Feng}
\ead{fengjinhua327@163.com}

\author[add1]{Bairong Shen\corref{cor1}\orcidlink{0000-0003-2899-1531}}
\ead{Bairong.shen@scu.edu.cn}

\cortext[cor1]{Corresponding author}

\affiliation[add1]{organization={Department of Ophthalmology and Institutes for Systems Genetics, Frontiers Science Center for Disease-related Molecular Network, West China Hospital},
            addressline={Sichuan University}, 
            city={Chengdu},
            postcode={610212}, 
            state={Sichuan},
            country={China}}

\begin{abstract}
     In recent years, Facial Expression Recognition (FER) has gained increasing attention. 
     Most current work focuses on supervised learning, which requires a large amount of labeled and diverse images, while FER suffers from the scarcity of large, diverse datasets and annotation difficulty.  
     To address these problems, we focus on utilizing large unlabeled Face Recognition (FR) datasets to boost semi-supervised FER. 
     Specifically, we first perform face reconstruction pre-training on large-scale facial images without annotations to learn features of facial geometry and expression regions, followed by two-stage fine-tuning on FER datasets with limited labels.
     In addition, to further alleviate the scarcity of labeled and diverse images, we propose a Mixup-based data augmentation strategy tailored for facial images, and the loss weights of real and virtual images are determined according to the intersection-over-union (IoU) of the faces in the two images. 
     Experiments on RAF-DB, AffectNet, and FERPlus show that our method outperforms existing semi-supervised FER methods and achieves new state-of-the-art performance.  
     Remarkably, with only 5$\%$, 25$\%$ training sets, our method achieves 64.02$\%$ on AffectNet, and 88.23$\%$ on RAF-DB, which is comparable to fully supervised state-of-the-art methods.
     Codes will be made publicly available at \href{https://github.com/zhelishisongjie/SSFER}{https://github.com/zhelishisongjie/SSFER}. 
\end{abstract}

\begin{highlights}
\item Face reconstruction pre-training learns facial features and expression regions.
\item Proposed augmentation method balances real and virtual image loss weights using IoU.
\item State-of-the-art performance achieved in semi-supervised setting.
\item Extensive experiments show effectiveness and generalizability across facial tasks.
\end{highlights}

\end{frontmatter}

\section{Introduction}
Facial expressions are one of the most common ways of expressing human emotions and play an important role in human communication. 
Facial Expression Recognition (FER) has great potential for application in the fields of assisted driving \cite{jeong2018driver,liu2020driver}, lecturing \cite{whitehill2014faces}, intelligent medical care \cite{jin2020diagnosing,yun2017social}, virtual reality \cite{chen2023emotion}, and so on.
Despite its promising applications, similar face domain tasks such as Face Recognition (FR) have become almost mature technology, whereas FER still faces developmental bottlenecks.
We observe two major challenges in FER that contribute to the significant gap between it and FR technologies.

\begin{figure}[htbp]
    \centering
    \includegraphics[width=4.0in]{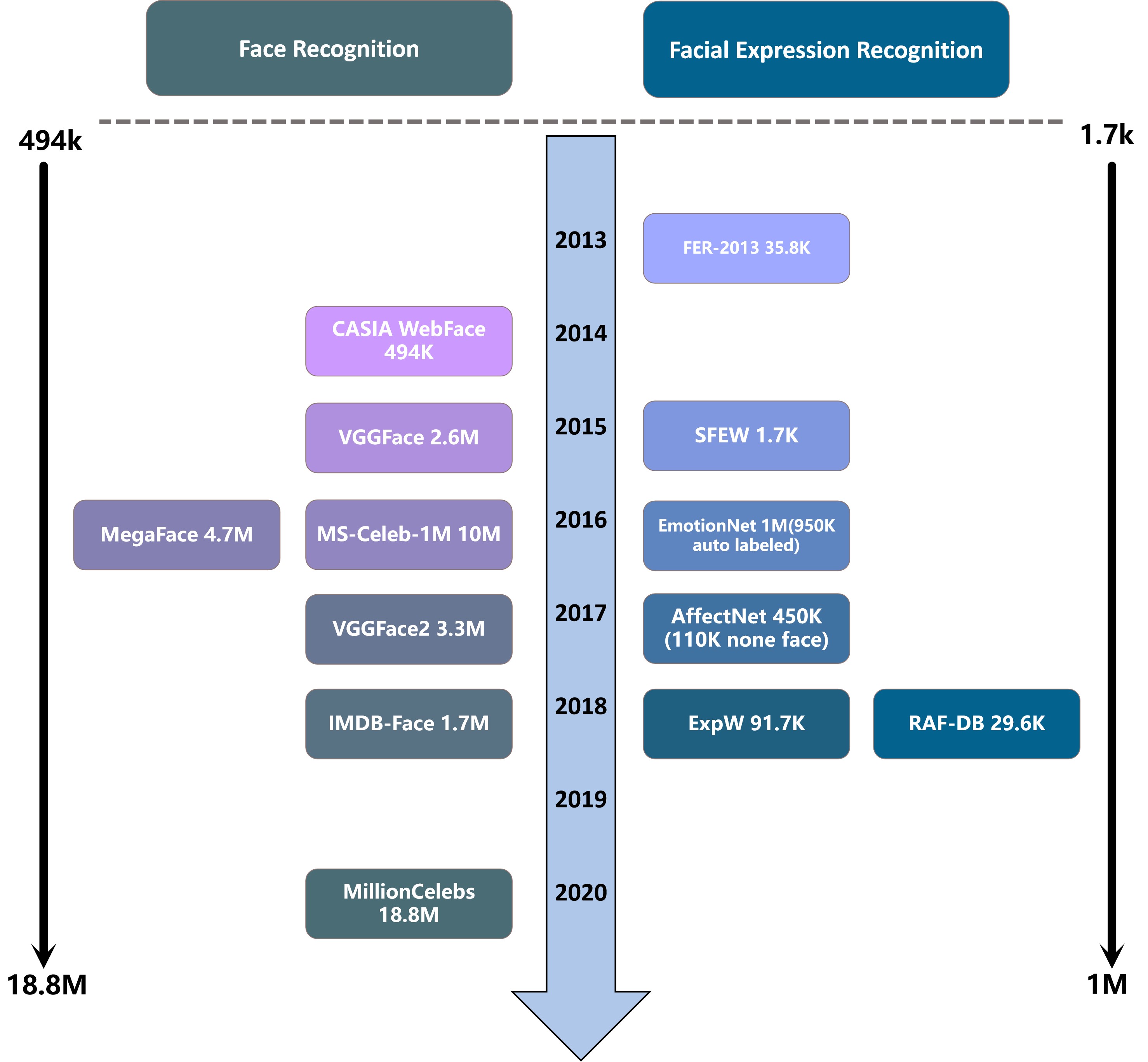}
    \caption{\textbf{The evolution of face recognition datasets and facial expression recognition datasets}}
    \label{fig FRvsFER}
\end{figure}

FER faces the \textbf{challenge of scarcity of large and diverse datasets. }
As illustrated in Fig. \ref{fig FRvsFER}, there is a significant disparity in the number of datasets available for FR and FER, making it challenging to train accurate and robust FER models. 
Additionally, FER faces the \textbf{challenge of annotation difficulty} since untrained persons annotating expression data is difficult and time-consuming.
Bartlett et al. \cite{bartlett1999measuring} showed that it takes more than 100 hours of training for a person to achieve 70$\%$ accuracy in recognizing facial movements associated with expressions. 
Meanwhile, Susskind et al. \cite{susskind2007human} showed that an experienced psychology student achieved an accuracy of 89.20$\%$ in an experiment with six expressions. 
In comparison, untrained persons achieved an accuracy of 99.20$\%$ \cite{kumar2009attribute} on the LFW \cite{LFW} face recognition dataset.

Therefore, researchers have turned to enhancing FER through transfer learning or other technologies \cite{facenet2expnet,Meta-Face2Exp} by leveraging the extensive available FR datasets. 
Although these works demonstrate their effectiveness, since the annotations in the face recognition dataset are different from those in the FER dataset, face identity information may impair the model and weaken the expression discrimination of the learned features \cite{DeepFER}. 
Based on the above observations, we seek to address these challenges to boost the semi-supervised FER by utilizing unlabeled FR datasets and avoiding using the identity information within them.


To better utilize unlabeled images, we propose employing semi-supervised learning and self-supervised learning as effective strategies to transfer learning from knowledge learned on large-scale FR datasets. 
While many works have proposed methods that combine semi-supervised and self-supervised learning with a multi-stage pipeline \cite{chen2020big,semivit, kim2021selfmatch, ke2022three, joe2022contracluster, tian2023few} but none has yet been applied to the FER task. 
Here, we adopt a similar three-stage Self-supervised Semi-supervised Facial Expression Recognition (SSFER) framework, with the key difference being the inclusion of face reconstruction pre-training aimed at learning general facial features. 
This encourages the model to understand invariant patterns and relationships, such as facial geometry and expression regions within the faces.
In addition, to further alleviate the scarcity of labeled and diverse images. 
We proposed FaceMix, a data augmentation method specifically designed for facial images. 
During each training session, the model is trained on both virtual and real images, and the loss weights of the real and virtual images are determined according to the intersection-over-union (IoU) of the faces in the two images.
This allows the model to add diverse training samples while ensuring that it is trained on high-quality samples.

In this paper, we proposed a hybrid data-efficient semi-supervised method to effectively utilize the information from unlabeled facial images. 
First self-supervised pre-training on large-scale unlabeled facial images, second supervised fine-tuning with FaceMix augmentation on labeled FER images, third semi-supervised fine-tuning on unlabeled FER images. 
Notably, the vanilla ViT-Base is used in our SSFER framework without modification. 
Overall, our contributions are as follows: 

\begin{itemize}
    \item{We perform face reconstruction pre-training on unlabeled facial images to learn general facial features and understand invariant patterns and relationships, such as facial geometry and expression regions. }
    \item{We proposed a data augmentation strategy, FaceMix, which is more suitable for facial images, taking into account facial angle and pose, allowing the model to add diverse training samples while ensuring that it is trained on high-quality samples. }
    \item{We provide a framework that is scalable and extensible to a variety of facial tasks, with additional experiments verifying its effectiveness and robustness. }
    \item{Extensive experiments were conducted on three benchmark datasets, and our method outperformed state-of-the-art methods. This sets strong benchmarks for future semi-supervised FER.}
\end{itemize}

\section{Related work}
\subsection{Facial Expression Recognition} 
In recent years, Facial Expression Recognition(FER) has gained increasing attention. 
Supervised learning approaches have made remarkable progress in FER with the development of deep learning. 
Wang et al. \cite{SCN} proposed a self-cure network (SCN) to reduce the uncertainty and avoid the network to overfit incorrectly labeled samples. 
Zhang et al. \cite{zhu2024emotion} proposed an emotion knowledge-based fine-grained (EK-FG) recognition network that leverages 135 fine-grained emotions and prior knowledge to effectively distinguish subtle facial expressions. 
Kim et al. \cite{kim2024towards} proposed FAAT, a facial attention-based adversarial training method to enhance the robustness of facial expression recognition models against test-time attacks, demonstrating significant improvements in model performance under adversarial perturbations. 

The examples of supervised learning have effectively extracted facial expression features with very promising performance.
However, they neglect the scarcity of data and are limited by the lack of labeled images.
A natural solution to this problem is semi-supervised learning.
Jiang et al. \cite{jiang2021boosting} automatically and progressively select clean labeled training images to reduce label noise. Use the collected clean labeled images to compute supervised classification loss and unlabeled images to compute unsupervised consistency loss.
Li et al. \cite{Ada-CM} proposed an Adaptive Confidence Margin (Ada-CM) that splits all unlabeled images in two to fully utilize all unlabeled images for semi-supervised FER by comparing the confidence scores of each training period with the adaptively learned confidence margin.

Despite the remarkable progress that has been made, examples of semi-supervised FER do not fully exploit the information in the images themselves.
In this work, We construct a large-scale facial images dataset for self-supervised pre-training followed by semi-supervised fine-tuning on the FER dataset, which greatly alleviates the constraint of data labeling.

\subsection{Self-Supervised Learning}

In recent years, the Masked Language Model (MLM) has become an efficient self-supervised strategy in Natural Language Processing (NLP).
Pre-trained models such as BERT \cite{bert} are designed to reconstruct masked tokens in a corpus.
Inspired by MLM, BEiT \cite{beit} presented a masked image modeling framework to pretrain vision transformers. 
The image patches are tokenized into visual tokens with DALL-E \cite{DALLE}. 
These tokens are then masked randomly before being fed into the transformer backbone, and the training goal is to reconstruct the original visual tokens from the corrupted patches.

MAE \cite{MAE} proposed a more direct approach to masked image modeling by directly reconstructing the pixels of masked patches, MAE is simpler and faster without a tokenizer, so we follow MAE as the pre-training framework.
However, MAE, BEiT, and other frameworks are pre-trained on various types of images.

To focus on facial visual feature learning, we construct a facial image dataset of 1.2 million facial images.
Including the existing FR and FER datasets such as CASIA-WebFace \cite{WebFace} and AffectNet \cite{AffectNet}. 
Explore the potential of self-supervised pre-training on large-scale facial images.

\subsection{Semi-Supervised Learning}
Deep learning-based FER requires a large number of labeled images, but building large datasets with high-quality labels is both time-consuming and labor-intensive.
Semi-supervised is another effective approach to address the lack of labeled images, FixMatch \cite{fixmatch}, known as a popular semi-supervised method in the last few years, can be interpreted as a teacher-student framework, in which the student model and the teacher model are identical.

In Fixmatch, weak augmented inputs and strong augmented inputs share the same model, which often results in a model that tends to collapse \cite{grill2020bootstrap}, so we used the exponential moving average (EMA)-teacher.

EMA-teacher is an updated version of FixMatch, where pseudo-labels are produced by the teacher model and the teacher parameters are determined by an exponential moving average (EMA) of the student parameters, EMA has been successful in many tasks, such as semi-supervised human action recognition \cite{xing2023svformer}, and speech recognition \cite{manohar2021kaizen,higuchi2021momentum}. 
Here we were the first to adopt it in semi-supervised FER.

\subsection{Vision Transformer}
Transformer \cite{transformer} has made significant progress in Natural Language Processing (NLP), Vision Transformers (ViT) \cite{ViT} introduces the transformer architecture to the vision domain.
ViT captures long-range dependencies among patches by splitting the image into patches, and then the self-attention mechanism in the transformer captures the dependencies between these patches.

Subsequently, some researchers have also introduced ViT into FER.
Aouayeb et al. \cite{aouayeb2021learning} applied the Vision Transformer (ViT) architecture to the FER task by incorporating the Squeeze-and-Excitation (SE) block prior to the Multi-Layer Perceptron head.
Zheng et al. \cite{poster} designed a transformer-based cross-fusion method to facilitate efficient cooperation between facial landmarks and image features, maximizing attention to salient facial regions.
Xue et al. \cite{transfer} proposed TransFER, which is a transformer-based approach that extracts attention information using multi-branch local CNNs and multi-head self-attention in ViT. The approach ensures diversity in attention features through a multi-attention-dropping module.
However, the above approaches focused on the fully supervised setting, and limited efforts have been made on vision transforms in the semi-supervised FER.

\subsection{Mixup}
Mixup \cite{mixup} is a simple and powerful data augmentation strategy for training deep learning-based on convex combinations of images and labels, which has a wide range of applications in computer vision, natural language processing, and audio tasks.
In FER, Mixup helps to alleviate the scarcity of large and diverse datasets \cite{mixaugment,marginmix}, but facial images vary greatly in head pose and angle.
Mixup can cause the mixed images to differ from the real images, potentially hindering the learning of the model.

To solve this problem, we propose a simple method called FaceMix. 
Specifically, during each training session, the model is trained on both virtual and real images, and the loss weights of the real and virtual images are determined according to the IoU of the faces in the two images. 
This approach allows the model to incorporate diverse training samples while ensuring that it is trained on high-quality samples.

\section{Method}
\subsection{Pipeline}
Many works have proposed methods that combine semi-supervised and self-supervised learning with a multi-stage pipeline \cite{chen2020big,semivit, kim2021selfmatch, ke2022three, joe2022contracluster, tian2023few} but none has yet been applied to the FER task. 
Here we are the first to adopt this paradigm to FER with the difference that we pre-trained the model on facial images using a reconstruction pre-training aimed at learning general facial features and encouraging the model to understand invariant patterns and relationships such as facial geometry and expression regions within the faces.
Furthermore, we applied FaceMix during the supervised stage to alleviate the scarcity of large and diverse datasets. 

\begin{figure*}[htbp]
    \centering
    \includegraphics[width=6.7in]{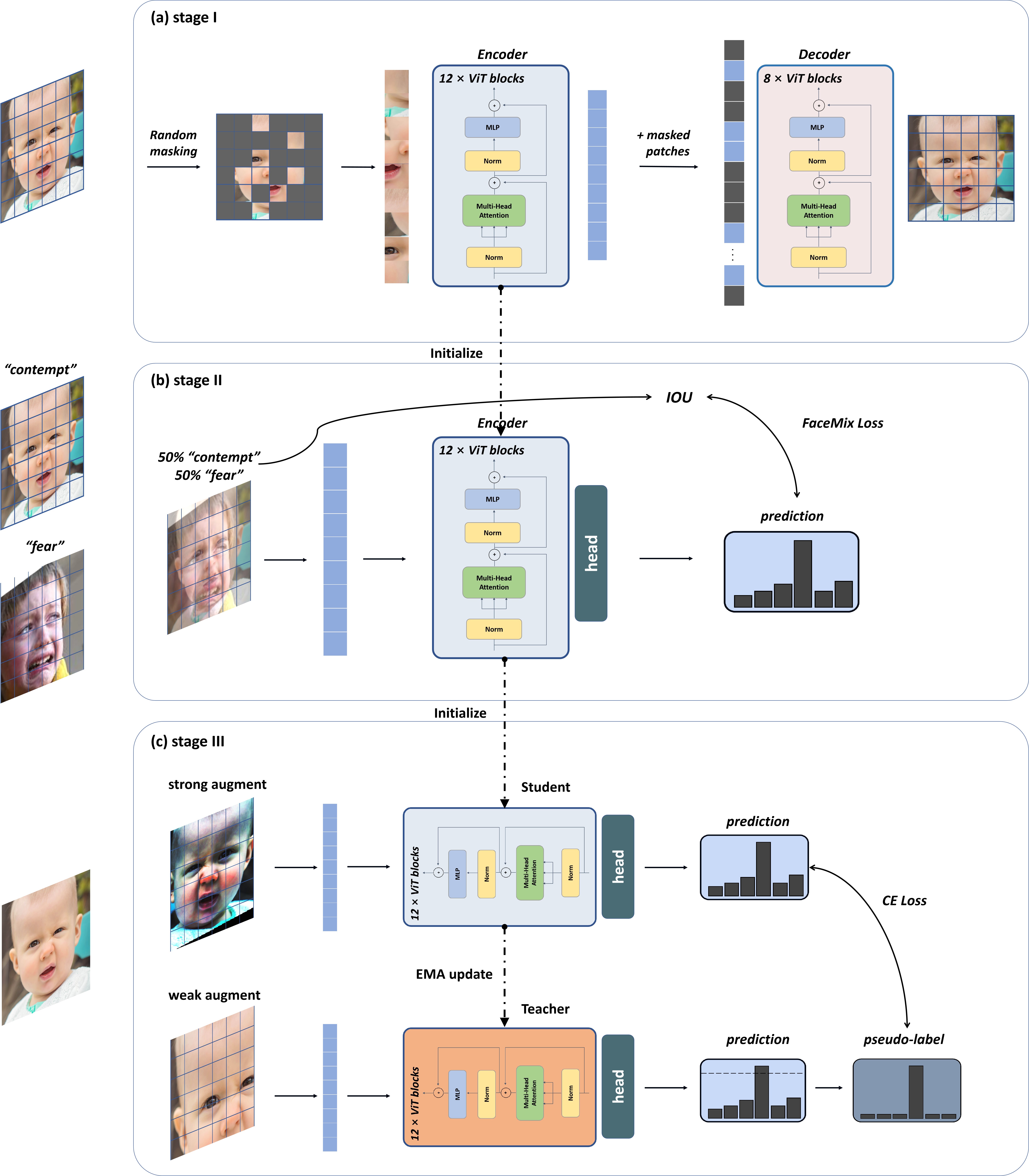}
    \caption{\textbf{The pipeline of our SSFER.} 
    It consists of three stages: 
    (a) \textbf{Self-Supervised Pre-training Stage}, unlabeled images are first divided into patches and then 75$\%$ of them are randomly masked, the remaining 25$\%$ of the visible patches are fed into the ViT Encoder, and the output embedding together with the masked patches are fed into the ViT decoder for image reconstruction; 
    (b) \textbf{Supervised Fine-tuning Stage}, convex combinations of images and labels are divided into patches and then fed into the ViT encoder and the MLP head, and then the FaceMix Loss is calculated by the predictions and the IOUs of the images; 
    (c) \textbf{Semi-Supervised Fine-tuning Stage}, unlabeled images are fed into the student model after strong augmentation and into the teacher model after weak augmentation, with the teacher parameters updated by an exponential moving average of the student model. If the prediction confidence of the teacher model is higher than the threshold, the class with the highest confidence is used as pseudo-labels to calculate the cross-entropy loss with the student predictions;
    \textbf{Notably}, the ViT encoder in our SSFER framework is the vanilla ViT-Base without modifications. 
    }
    \label{fig pipeline}
\end{figure*}

The SSFER training pipeline is shown in Fig. \ref{fig pipeline}. 
First, self-supervised pre-training is performed on large unlabeled facial images.
Then, standard supervised fine-tuning is conducted on the available labeled images. 
Finally, semi-supervised fine-tuning is performed on both labeled and unlabeled images.
\textbf{The mathematical pseudo-codes for the whole pipeline are provided in the supplementary Algorithm S1, S2, and S3.}

\subsection{Self-Supervised Pre-training}
\begin{figure}[htbp]
    \centering
    \includegraphics[width=4.0in]{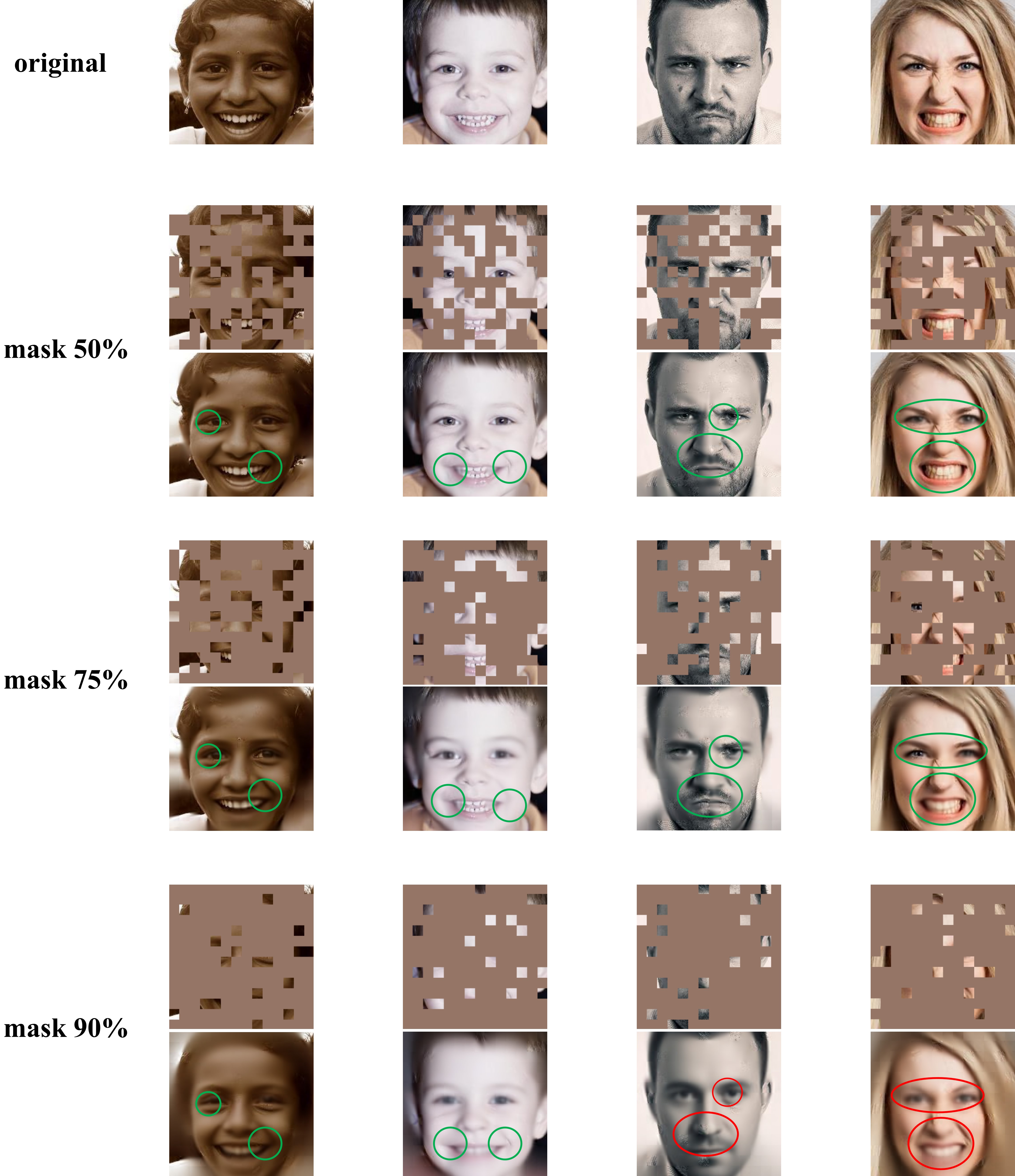}
    \caption{\textbf{Samples of different masking ratios. }
    Red circles represent failure to reconstruct expression regions, green circles represent success to reconstruct expression regions. }
    \label{fig different mask ratio}

\end{figure}

Expression regions refer to those facial regions associated with expressions (e.g., eyes, mouth), and in the self-supervised pre-training stage we focus on reconstructing these key regions and facial geometric relationships. 
We train vanilla ViT as Masked Autoencoders \cite{MAE} on about 1.2 million unlabeled facial images, we split the input image into patches and randomly mask a number of them, leaving visible patches to the encoder.
In the reconstruction process, all the patches are fed into the decoder for pixel-level reconstruction.
\textbf{Samples of different masking ratios are shown in Fig. \ref{fig different mask ratio}. }

When using a mask ratio of 90$\%$, the reconstruction of the expression regions fails, and the angry samples tend to be reconstructed as neutral. 
Therefore, we use a 75$\%$ mask ratio, which provides better reconstruction results and speeds up the pre-training stage, as the MAE encoder only needs to process 25$\%$ of the patches.

\subsection{Semi-Supervised Fine-tuning}
In EMA-teacher, the teacher parameters are updated using an exponential moving average of the student parameters, which can be expressed as follows:
\begin{equation}
    \label{equ EMA params}
    \boldsymbol{\theta_t} \leftarrow m\boldsymbol{\theta_t} + (1-m)\boldsymbol{\theta_s}
\end{equation}
where \(m\) represents the momentum coefficient, and \(\boldsymbol{\theta_t}\) and \(\boldsymbol{\theta_s}\) are the parameters of the teacher and student models, respectively.
For labeled samples $(\boldsymbol{x^l}, \boldsymbol{y^l})$, the loss is the standard cross-entropy loss, which can be expressed as follows: 
\begin{equation}
    \label{equ labeled loss}
    \mathcal{L}_l = -\frac1N_L \sum^{N_L} CE(\boldsymbol{x^l}, \boldsymbol{y^l})
\end{equation}

For unlabeled samples \(\boldsymbol{x^u}\), weak and strong augmentation are used to obtain \(\boldsymbol{\mathcal{A}_{weak}(x^l)}\) and \(\boldsymbol{\mathcal{A}_{strong}(x^u)}\). 
The weakly augmented inputs are fed into the teacher model and output the classes probabilities $\boldsymbol{p}$, then select the largest probabilities from $\boldsymbol{p}$ and their corresponding classes as the pseudo-labels $\boldsymbol{\hat{y}}$, the process can be expressed as follows: 


\begin{equation}
    \label{equ pseudo label}
    \boldsymbol{\hat{y}} = \arg\max_{c} f(\boldsymbol{\mathcal{A}_{weak}(x^u)};\boldsymbol{\theta_t})_c
\end{equation}

If the pseudo-label confidence is higher than the predefined confidence threshold $\tau$, then $\boldsymbol{\hat{y}}$ will be used as a supervisory signal to supervise student learning on the unlabeled strong augmented samples \(\boldsymbol{\mathcal{A}_{strong}(x^u)}\).
The loss of unlabeled samples can be expressed as follows: 
\begin{equation}
    \label{equ unlabeled loss}
    \mathcal{L}_u = -\frac1N_U \sum^{N_U} \mathbb{I}(\max(\boldsymbol{p})>\tau)CE(\boldsymbol{\mathcal{A}_{strong}(x^u)}, \boldsymbol{\hat{y}})
\end{equation}
where $\mathbb{I}$ is the indicator function, which is 1 if $(\max(\boldsymbol{p}) > \tau)$ and 0 otherwise. The overall loss function is given by: 
\begin{equation}
    \label{equ overall loss}
    \mathcal{L} = \mathcal{L}_l + \mu\mathcal{L}_u
\end{equation}
where $\mu$ is the trade-off weight between the labeled and unlabeled losses.

\subsection{FaceMix}
An example of a convex combination of pairs of samples and their labels by Mixup \cite{mixup} to construct a virtual sample can be expressed as follows:  
\begin{equation}
    \begin{aligned}
        \label{equ mix}
        \boldsymbol{\tilde{x}} &= \lambda \boldsymbol{x_i} + (1-\lambda) \boldsymbol{x_j} \\
        \boldsymbol{\tilde{y}} &= \lambda \boldsymbol{y_i} + (1-\lambda) \boldsymbol{y_j}
    \end{aligned}
\end{equation}
Where, $(\boldsymbol{\tilde{x}},\boldsymbol{\tilde{y}})$ are the virtual samples after mixing, the ratio $\lambda$ is a scalar conforming to the beta distribution, $\boldsymbol{x_i}$ and $\boldsymbol{x_j}$ are the input images, and $\boldsymbol{y_i}$ and $\boldsymbol{y_j}$ are their corresponding one-hot labels.

The training implementation of Mixup is very simple, with minimal computational cost.
When training with Mixup, the soft target cross-entropy loss used for the virtual samples can be expressed as follows:
\begin{equation}
    \label{equ virtual loss}
    \mathcal{L}_v = -\frac1N \sum^{N} \boldsymbol{\tilde{y}} \cdot \log\boldsymbol{\hat{p}}
\end{equation}
where $\boldsymbol{\hat{p}}$ is the prediction probabilities of the samples $\boldsymbol{\tilde{x}}$, and $(\boldsymbol{\tilde{x}}, \boldsymbol{\tilde{y}})$ are given in (\ref{equ mix}).

\begin{figure}[htbp]
    \centering
    \includegraphics[width=3.4in]{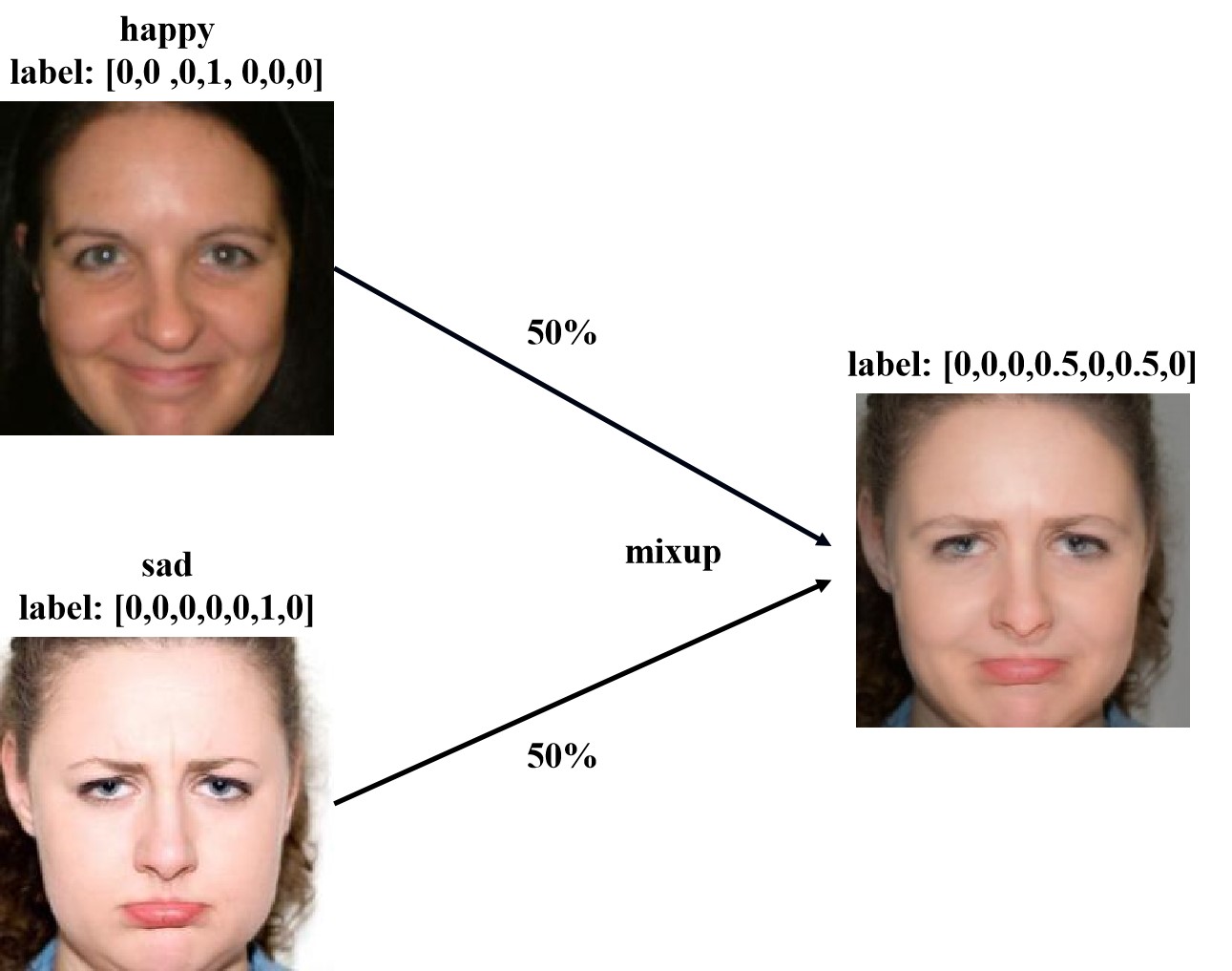}
    \caption{\textbf{Use Mixup to mix two facial images to construct a virtual sample. }}
    \label{fig mixup one}
\end{figure}

An example of Mixup constructing a virtual sample is shown in Fig. \ref{fig mixup one}, two original face images are linearly mixed in the ratio 50:50, the generated face combines the facial features of the two original images, and its label states that this virtual sample has 50$\%$ of "happy" and 50$\%$ of "sad" classes. 

\begin{figure}[htbp]
    \centering
    \includegraphics[width=3.4in]{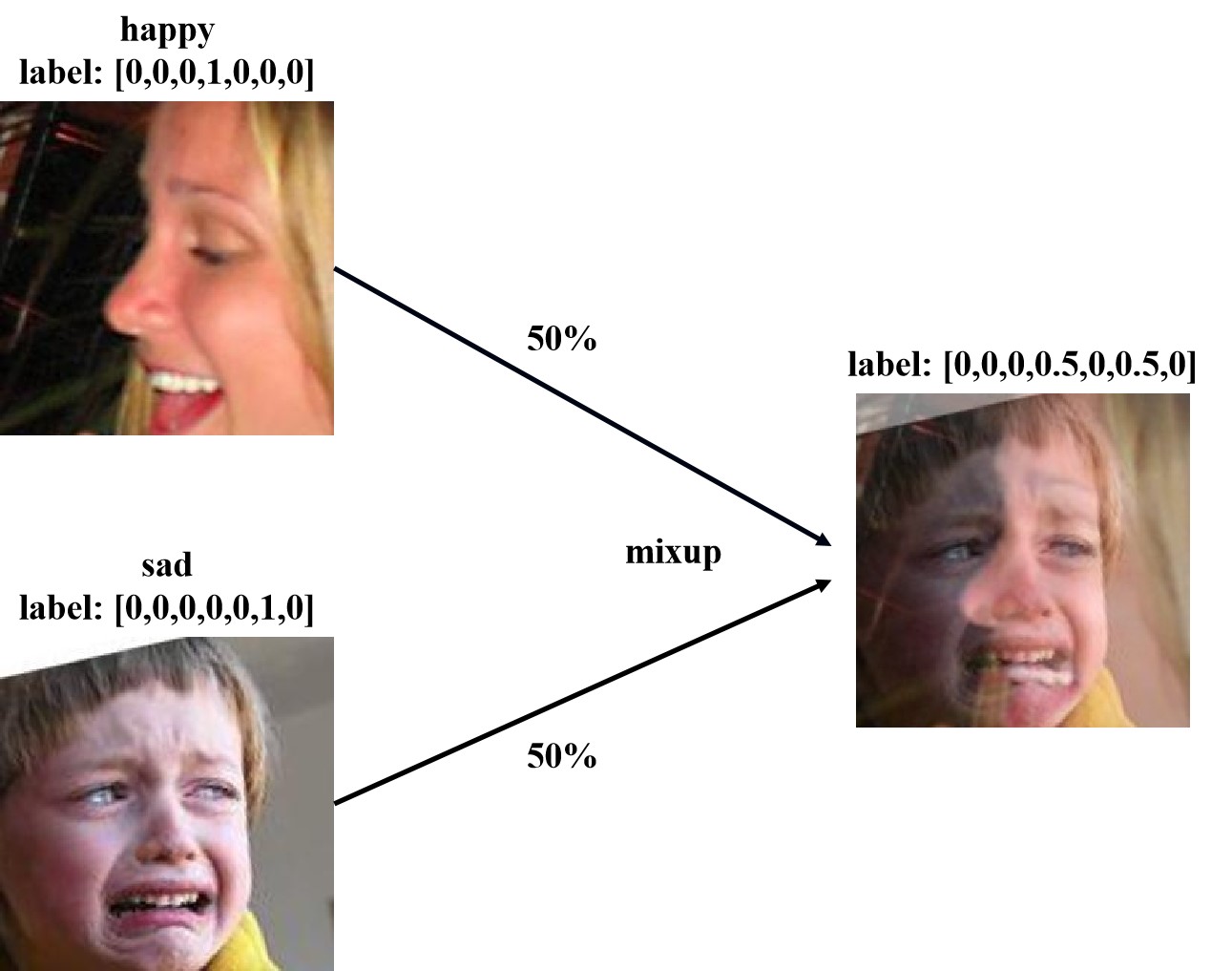}
    \caption{\textbf{Example of mixing images with different head angles. }}
    \label{fig mixup two}
\end{figure}

But when the head pose and angle of the two original face images are very different, the constructed virtual samples are different from the real face as shown in Fig. \ref{fig mixup two}. This situation may hinder the training and learning of the model.

To address this problem, we propose an improved Mixup-based method, called FaceMix.
First, we use yolo-face to detect the face boxes, and calculate the $IOU$ of the two faces to get a coefficient $\kappa$, this process can be expressed as follows: 
\begin{equation}
    \label{equ iou compute}
    IoU=\frac{\text{Area of overlap } B_i \text{ and } B_j} {\text{Area of union } B_i \text{ and } B_j}
\end{equation}

\begin{equation}
    \label{equ kappa}
    \kappa=1-IOU
\end{equation}
Where $B_i$ and $B_j$ are the face boxes of $\boldsymbol{x_i}$ and $\boldsymbol{x_j}$. 
Moreover, we also tried other metrics besides IOU such as Peak Signal-to-Noise Ratio (PSNR), Structural Similarity Index (SSIM) \cite{SSIM}, and Feature Similarity Index (FSIM) \cite{FSIM} to calculate $\kappa$. 
During each training session, the model is trained on both virtual and real samples, and the FaceMix loss can be expressed as follows: 
\begin{equation}
    \label{equ FaceMix loss}
    \mathcal{L}=\mathcal{L}_v+\kappa(\mathcal{L}_i+\mathcal{L}_j)
\end{equation}

Where $\mathcal{L}$, $\mathcal{L}_v$ are given in (\ref{equ virtual loss}), $\kappa$ is given in (\ref{equ kappa}), 
$\mathcal{L}_i$ and $\mathcal{L}_j$ are the standard categories cross-entropy loss of two random samples $\boldsymbol{x_i}$ and $\boldsymbol{x_j}$, respectively, which are described in (\ref{equ mix}).

\begin{figure}[htbp]
    \centering
    \includegraphics[width=3.4in]{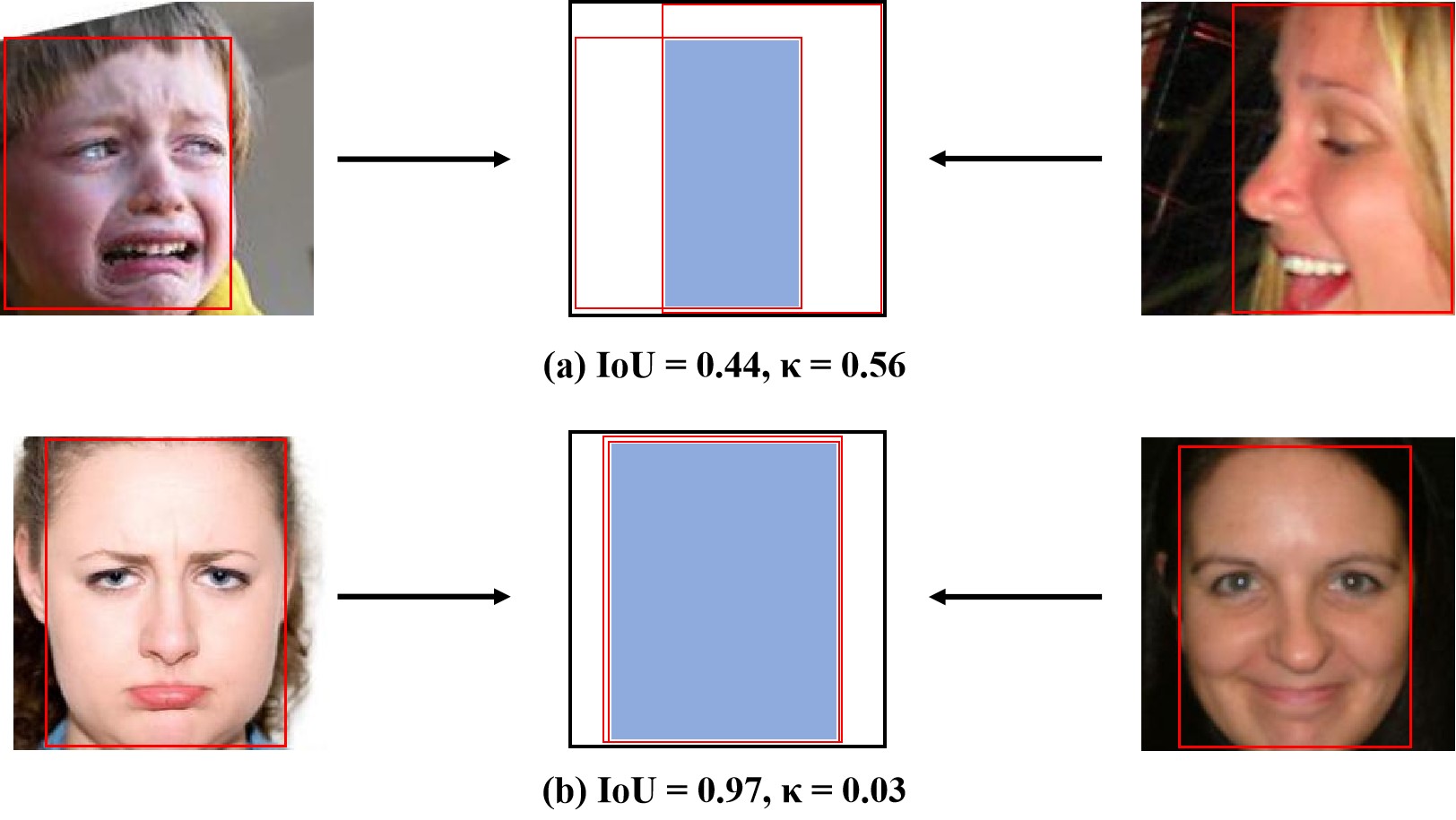}
    \caption{\textbf{Examples of IoU calculation. }}
    \label{fig IoU compute}
\end{figure}

An example of IoU computing is shown in Fig. \ref{fig IoU compute}. 
When the angle and head pose of two faces tend to be the same (Fig. \ref{fig IoU compute} (b)), $\kappa$ will be close to 0, at this time the model focuses on the training of the high-quality virtual samples; 
Conversely, when the difference between the two faces is significant (Fig. \ref{fig IoU compute} (a)), $\kappa$ increases, the mixing loss will be merged with the classification loss, to enhance the ability to classify the original samples and the virtual samples. 
This allows the model to increase the number of training samples while ensuring that it is trained on high-quality samples. 

\section{Experiments}
\subsection{Datasets}

RAF-DB \cite{RAFDB} is a facial expression dataset containing about 30,000 various facial images retrieved from the Internet.
Each image has been independently annotated by approximately 40 annotators based on crowdsourced annotation.
RAF-DB is composed of two subsets: a single-label subset, which includes images annotated with seven basic expressions; and a multi-label subset, which includes images annotated with 12 composite expressions.
For our experiments, we used the single-label subset, which contains 12,271 training images and 3,068 test images.

AffectNet \cite{AffectNet} is a dataset of 1 million facial images collected from the Internet. About 450,000 of these images are labeled and were collected using 1250 emotion-related keywords in six different languages. It is currently the largest dataset available in FER. 
The 7-class setting contains the basic expression categories, comprising 283,901 training images and 3500 validation images. 
The 8-class setting contains the seven basic expression categories and contempt, comprising 287,651 training images and 4000 validation images.
For our experiment, we used both the 7-class and 8-class settings.

FERPlus \cite{FERPlus} is an extended version of FER2013, collected by the Google search engine, where each image is labeled by 10 annotators to 8 sentiment categories (the seven basic expression and contempt), and the seven basic expressions and contempts. 
FERPlus includes 28,709 training images, 3589 validation images, and 3573 test grayscale images. 
We merged the FERPlus validation set with the test set, which was not used during training.
The detailed dataset configuration is shown in Table \ref{tab dataset}.
\begin{table}[htbp]                    
    \caption{\textbf{Detailed Size of The Experimental Dataset.}}        
        \label{tab dataset}
    \centering      
    \resizebox{0.55\textwidth}{!}
    {                       
    \begin{tabular}{cccc}               
    \toprule                            
        Dataset&Training set size&Testing set size&Classes \\
    \midrule                            
        RAB-DB              &12271  &3068 &7 \\
        AffectNet(7-class)  &287401 &3500 &7 \\
        AffectNet(8-class)  &291651 &4000 &8 \\
        FERPlus             &28709  &7178 &8\\
    \bottomrule                         
    \end{tabular}
    }
\end{table}

\subsection{Implementation Details}
As a regular practice, we first detected and aligned all the images, and resized them to 224$\times$224 pixels.

In the self-supervised pre-training stage, we construct a facial image dataset of 1.2 million facial images, including the existing FR and FER datasets such as CASI-WebFace \cite{WebFace} and AffectNet \cite{AffectNet}.
We pre-train ViT-Base for 600 epochs with 50 warm-up epochs. 
The learning rate is 3.4e-4, batch size is set to 256.
other configuration we follow He et al. \cite{MAE} to reconstruct the normalized pixel values of each masked patch.

In the supervised fine-tuning stage, we fine-tune 100 epochs with 5 warm-up epochs.
For experiments with less than 500 labeled images, the number of training epochs is expanded to 1000 epochs.
The learning rate is 1.0e-4, the minimum learning rate is set to 1e-5, and the learning rate for warm-up initialization is set to 5e-5. 
Batch size is set to 32.
FaceMix is applied in the supervised fine-tuning stage. 

In the semi-supervised fine-tuning stage, we fine-tune 50 epochs.
Weak augmentations include random resize crop and random horizon flip.
Strong augmentations include random resize crop, random horizon flip, and randaugment. 
The learning rate is 1.5e-4, batch size is set to 64.
Our framework is trained on 24 DCUs (performance similar to V100).

\subsection{Results Visualisation}
\begin{figure*}[htbp]
    \centering
    \includegraphics[width=6.7in]{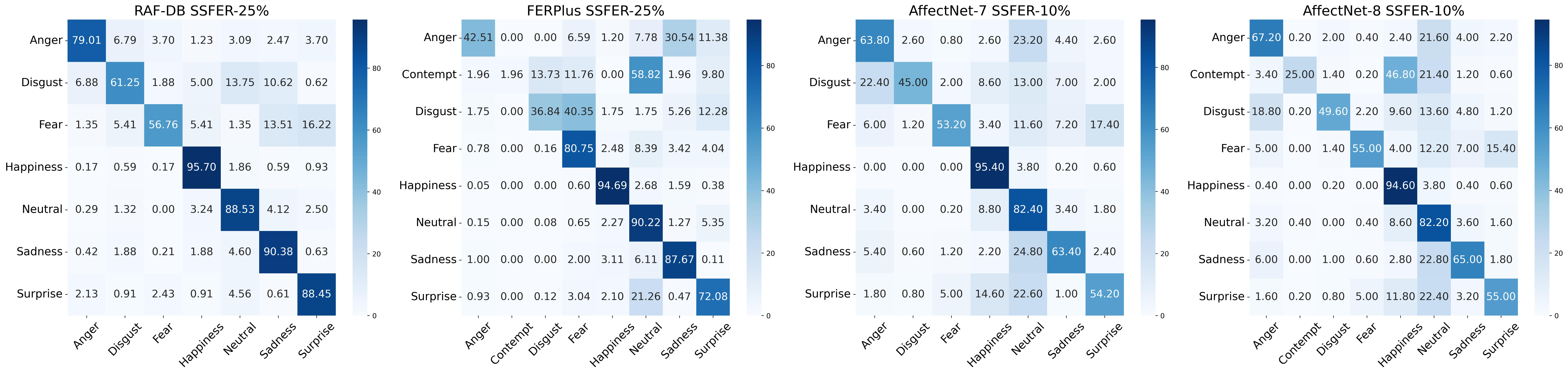}
    \caption{The confusion matrices of our SSFER on RAF-DB, FERPlus and AffectNet. }
    \label{fig confusion matrix}
\end{figure*}

We show the confusion matrices of our SSFER in Fig. \ref{fig confusion matrix}, and we find that the accuracy in the categories 'Neutral', 'Happiness', and 'Sadness' is very high with only 25$\%$ of the labeled images, reaching 95.70$\%$, 88.53$\%$ and 90.38$\%$ on RAF-DB, respectively, outperforming existing state-of-the-art methods (e.g, the highest accuracies reported by DCJT \cite{DCJT} on RAF-DB are 87.35$\%$, 93.92$\%$, 84.31$\%$ respectively). 
This can be explained by the fact that these three expressions ('Neutral', 'Happiness', and 'Sadness') discriminators occur in regular expression regions: stable mouth corners for 'Neutral', upturned mouth corners for 'Happiness', and downturned mouth corners for 'Sadness'. 
In the pre-training reconstruction, our model is able to learn the detailed features of these regions to reconstruct the image completely, as shown in Fig. \ref{fig different mask ratio}.

For FERPlus, 'Contempt', 'Disgust' perform particularly poorly, mainly because their training images are only 165 and 191 respectively, and we only use 25$\%$ of the images. 
More importantly, FERPlus has only 30 and 21 test images respectively, which is very unbalanced compared to the other classes, making it a more difficult challenge for semi-supervised algorithms.
\textbf{The SSFER confusion matrices for all ratios are shown in the supplementary Fig. S1}. 

\subsection{Comparison with fully supervised methods}
SSFER was also compared with fully supervised state-of-the-art methods, and the results are presented in Table \ref{tab sota compare}.
The performance of SSFER is comparable to the state-of-the-art methods on AffectNet when trained with 10$\%$ of the labels, and SSFER is also able to compete with the state-of-the-art methods on RAF-DB when trained with 25$\%$ of the labels.

Although SSFER fails in accuracy compared to fully supervised methods, its main advantage is its ability to achieve promising results with a small number of labeled samples.
Therefore, SSFER is more suitable for comparison with other semi-supervised FER methods.
\begin{table}[htbp]                    
    \centering
    \resizebox{0.50\textwidth}{!}{    
    \begin{threeparttable}
        \caption{\textbf{Comparison with Fully-Supervised State-of-The-Art Methods on RAF-DB, AffectNet, and FERPlus.}}        
        \label{tab sota compare}
                     
            \begin{tabular}{ccccc}               
            \toprule                            
                Method & RAF-DB & FERPlus & AffectNet-7 & AffectNet-8 \\
            \midrule   
                Multi-task EfficientNet-B2 \cite{savchenko2022classifying}   & - & - & 66.34 & 63.03 \\ 
                CAGE\cite{CAGE}                      & - & - & 66.60 & 62.30 \\
                LDL-ALSG \cite{LDL-ALSG}             & 85.53 & -     & 59.35 & -    \\      
                RAN \cite{RAN}                       & 86.90 & 89.16 & -     & 59.50    \\                   
                SCN \cite{SCN}                       & 87.03 & 89.35 & -     & 60.23    \\
                DACL \cite{DACL}                     & 87.78 & -     & 65.20 & -    \\
                KTN \cite{KTN}                       & 88.07 & 90.49 & 63.97 & -    \\
                EfficientFace \cite{EfficientFace}   & 88.36 & -     & 63.70 & 60.23 \\
                MA-Net \cite{MA-Net}                 & 88.42 & -     & 64.53 & 60.29 \\
                Meta-Face2Exp \cite{Meta-Face2Exp}   & 88.54 & -     & 64.23 & -    \\
                FENN \cite{FENN}                     & 88.91 & 89.53 & -     & 60.83    \\
                PSR \cite{PSR}                       & 88.98 & 89.75 & 63.77 & 60.68 \\
                EAC \cite{EAC}                       & 88.99 & 89.64 & 65.32 & -    \\
                AMP-Net \cite{AMP-Net}               & 89.25 & -     & 64.54 & 61.74 \\
                DMUE \cite{DMUE}                     & 89.42 & 89.51 & -     & 63.11 \\
                FDRL \cite{FDRL}                     & 89.47 & -     & -     & -        \\
                PT \cite{jiang2021boosting}          & 89.57 & 86.60 & -     & 58.54 \\
                DAN \cite{DAN}                       & 89.70 & -     & 65.69 & 62.09 \\
                MTAC \cite{MTAC}                     & 90.52 & 90.42 & -     & 62.28 \\
                TAN \cite{TAN}                       & 90.87 & 91.00 & 66.45 & - \\
                TransFER \cite{transfer}             & 90.91 & 90.83 & 66.23 & - \\ 
                DDAMFN \cite{DDAMFN}                 & 91.35 & 90.97 & 67.03 & 64.25 \\
                Tao et al. \cite{tao2024hierarchical}& 91.92 & -     & 66.97 & 63.82 \\
                APViT \cite{APViT}                   & 91.98 & 90.86 & 66.91 & - \\
                POSTER \cite{poster}                 & 92.05 & -     & 67.31 & 63.34 \\
                POSTER++ \cite{poster++}             & 92.21 & -     & 67.49 & 63.77 \\
                DCJT \cite{DCJT}                     & 92.24 & 88.95 & -     & -     \\
                
            \midrule 
                SSFER-1$\%$ (ours)   & - & - & 59.08 & 53.85  \\
                SSFER-5$\%$ (ours)   & 80.96 & 81.77 & 64.02 & 60.17  \\
                SSFER-10$\%$ (ours)  & 85.07 & 83.95 & 65.37 & 61.65  \\
                SSFER-25$\%$ (ours)  & 88.23 & 85.82 & - & -      \\
            \bottomrule                         
            \end{tabular}

    \end{threeparttable}
    }
\end{table}

\subsection{Comparison with semi-supervised methods}
\footnote{For convenient comparisons, we directly adopt the setup and results from \cite{roy2024exploring} and \cite{fang2023rethinking}. }
\textbf{Comparison with general semi-supervised methods }are shown in the Table \ref{tab semi compare}.
FixMatch achieves the highest accuracy of 63.25$\%$ on RAF-DB with 10 labels per class, but SSFER surpasses it as the number of labels increases. Our SSFER outperforms all other general semi-supervised methods.
\begin{table*}[htbp]                  
    \caption{\textbf{Comparison with General Semi-Supervised Methods on RAF-DB and AffectNet with Only 10, 25, 100, and 250 Labeled Images Per Class, a Total of 70, 175, 700, and 1750 Labeled Images for Training. }
    }        
    \label{tab semi compare}
    \centering
    \resizebox{0.90\textwidth}{!}
    {                     
    \begin{tabular}{c cccc cccc}               
    \toprule
    \multirow{2}{*}{\textbf{Method}}    &\multicolumn{4}{c}{\textbf{RAF-DB}}   &\multicolumn{4}{c}{\textbf{AffectNet-7}}     \\                   
    \cline{2-9}    
        &\textbf{10 labels} &\textbf{25 labels} &\textbf{100 labels} &\textbf{250 labels} &\textbf{10 labels}  &\textbf{25 labels} &\textbf{100 labels} &\textbf{250 labels}\\
    \midrule                            
        $\pi$-model \cite{pimodel}           &39.86&50.97&63.98&71.15&24.17&25.37&31.24&32.40 \\
        Pseudo-Label \cite{Pseudo-Label}     &58.31&39.11&54.07&67.40&18.00&21.05&33.05&37.37 \\
        Mean Teacher \cite{meanteacher}      &62.05&45.17&45.57&76.85&19.54&20.21&20.80&44.05 \\
        VAT \cite{VAT}                       &63.10&45.82&62.05&59.45&17.68&35.02&37.68&37.92 \\
        UDA \cite{UDA}                       &46.87&53.15&58.86&60.82&27.42&32.16&37.25&37.64 \\
        MixMatch \cite{mixmatch}             &36.34&43.12&64.14&73.66&30.80&32.40&39.77&48.31 \\
        ReMixMatch \cite{remixmatch}         &37.35&42.56&42.86&61.70&29.28&33.54&41.60&46.51 \\       
        FixMatch \cite{fixmatch}             &\textbf{63.25}&52.44&64.34&75.51&30.08&38.31&46.37&51.25 \\
        FlexMatch \cite{flexmatch}           &40.51&42.67&50.75&61.70&17.20&19.80&22.34&29.83 \\
        CoMatch \cite{comatch}               &40.04&52.59&68.05&73.46&21.23&23.54&27.45&30.31 \\
        CCSSL \cite{CCSSL}                   &50.59&51.30&63.79&74.93&16.89&21.34&24.46&28.94 \\
    \midrule  
        SSFER(ours)             &47.26&\textbf{59.81}&\textbf{75.29}&\textbf{83.11}&\textbf{32.85}&\textbf{42.94}&\textbf{52.71}&\textbf{57.42} \\
    \bottomrule                         
    \end{tabular}
    }
\end{table*}

\textbf{Comparison with state-of-the-art semi-supervised FER methods }are shown in the Table \ref{tab semi sota compare}. 
Our SSFER outperforms all existing state-of-the-art methods, demonstrating its superior performance in semi-supervised FER. 
It is important to note that the number of methods in this comparison is limited due to the relatively small amount of research on semi-supervised FER. 
To the best of our knowledge, the methods presented here represent the comprehensive set of semi-supervised FER methods currently available in the literature. 
\begin{table}[htbp]
\large                    
    \caption{\textbf{Comparison with State-of-The-Art Semi-Supervised Methods on RAF-DB and FERPlus Using Only 400, 1000 and 4000 Labeled Images for Training.}
    }        
        \label{tab semi sota compare}
    \centering
    \resizebox{0.55\textwidth}{!}
    {               
    \begin{tabular}{c cccccc c}               
    \toprule
    \multirow{2}{*}{\textbf{Method}}    &\multicolumn{3}{c}{\textbf{RAF-DB labels}}    &\multicolumn{3}{c}{\textbf{FERPlus(7-cls) labels}}   &\multirow{2}{*}{\textbf{Backbone}}\\                           
    \cline{2-7}
        &\multirow{1}{*}{\textbf{400}}   &\multirow{1}{*}{\textbf{1000}} &\multirow{1}{*}{\textbf{4000}}  &\multirow{1}{*}{\textbf{400}} &\multirow{1}{*}{\textbf{1000}} &\multirow{1}{*}{\textbf{4000}}\\
    \midrule                           
        MarginMix \cite{marginmix}                       &45.75&66.47&70.68   &56.75&59.38&75.18&  \multirow{5}{*}{WideResNet-28-2} \\
        Ada-CM \cite{Ada-CM}                             &59.03&68.38&75.98   &55.11&62.03&79.49&\\
        Progressive Teacher \cite{jiang2021boosting}     &51.54&67.35&71.29   &53.60&59.57&77.49&\\
        Meta-Face2Exp \cite{Meta-Face2Exp}                    &58.47&70.84&76.45   &55.69&64.34&80.52&\\
        Rethink-Self-SSL \cite{fang2023rethinking}       &62.36&72.92&77.41   &57.16&65.38&83.56&\\
    \midrule 
        MarginMix \cite{marginmix}                       &59.13&77.65&79.84   &54.73&62.91&74.12&  \multirow{5}{*}{ResNet-18-2} \\
        Ada-CM \cite{Ada-CM}                             &74.44&80.07&84.42   &55.46&63.11&74.96&\\
        Progressive Teacher \cite{jiang2021boosting}     &64.98&79.64&80.34   &52.30&61.98&73.38&\\
        Meta-Face2Exp \cite{Meta-Face2Exp}                    &72.17&81.22&84.65   &55.34&65.49&76.87&\\
        Rethink-Self-SSL \cite{fang2023rethinking}       &75.09&82.06&85.54   &58.87&65.48&75.91&\\
    \midrule
        SSFER(ours)                                     &\textbf{78.81}&\textbf{84.58}&\textbf{88.98}   &\textbf{77.61}&\textbf{81.15}&\textbf{85.62}&  ViT-Base \\
    \bottomrule                         
    \end{tabular}
    }
\end{table}

\subsection{Comparison of K-fold Validation}
To assess the effectiveness and reliability of SSFER, we conducted K-fold cross-validation on the RAF-DB and FERPlus training sets. 
Specifically, the dataset was divided into K equal-sized subsets. 
In each iteration, one subset served as the validation set, while the remaining K-1 subsets were used for training.
The results are shown in Table \ref{tab Kfold}, and our SSFER performs similarly to the test set in the K-fold setting.
Notably, the SSFER ViT-Base, trained with only 25$\%$ of the labels, outperforms the vanilla ViT-Large model that utilizes 100$\%$ of the labels. 
The slight drops in accuracy observed can be attributed to training the model with only 80$\%$ of the training set. 
\begin{table}[htbp]
    \caption{\textbf{K-fold Validation on RAF-DB and FERPlus.}
    K=5 was selected for comparison with other baseline models. }
    \label{tab Kfold}
    \centering
    \resizebox{0.55\textwidth}{!}{
    \begin{tabular}{c c c c c c c c c}               
    \toprule                           
    \textbf{Dataset} & \textbf{Method} & \textbf{Labels} & \textbf{Fold 1} & \textbf{Fold 2} & \textbf{Fold 3} & \textbf{Fold 4} & \textbf{Fold 5} & \textbf{Average}\\
    \midrule                            
    \multirow{8}{*}{RAF-DB} 
        & ViT-Base   & 9817  & 85.74 & 86.02 & 84.92 & 85.94 & 83.86 & 85.30 \\
        & ViT-Large  & 9817  & 86.43 & 87.28 & 86.39 & \textbf{86.52} & 85.90 & 86.50 \\
        & ResNet-50  & 9817  & 75.60 & 75.56 & 75.35 & 74.25 & 76.32 & 75.42 \\
        & VGG13      & 9817  & 82.15 & 81.90 & 82.80 & 82.64 & 81.95 & 82.29 \\
        & MobileNetV3 & 9817  & 77.67 & 77.95 & 76.89 & 77.13 & 77.63 & 77.45 \\
        \cmidrule(lr){2-9}
        & SSFER-5\%  & 491   & 77.25 & 78.68 & 76.86 & 76.79 & 77.78 & 77.47 \\
        & SSFER-10\% & 982   & 84.13 & 84.84 & 84.91 & 83.90 & 83.21 & 84.20 \\
        & SSFER-25\% & 2454  & \textbf{86.80} & \textbf{87.84} & \textbf{87.78} & 86.08 & \textbf{86.44} & \textbf{86.99} \\
    \midrule
    \multirow{8}{*}{FERPlus} 
        & ViT-Base   & 22967  & 84.68 & 85.06 & 85.10 & 85.45 & 85.40 & 85.14 \\
        & ViT-Large  & 22967  & \textbf{84.80} & 85.57 & 85.09 & 85.41 & \textbf{85.66} & \textbf{85.31} \\
        & ResNet-50  & 22967  & 79.69 & 80.78 & 79.97 & 79.55 & 79.94 & 79.99 \\
        & VGG13      & 22967  & 81.59 & 81.94 & 81.28 & 82.13 & 81.38 & 81.66 \\
        & MobileNetV3& 22967  & 80.59 & 81.15 & 80.59 & 80.34 & 80.52 & 80.64 \\
        \cmidrule(lr){2-9}
        & SSFER-5\%  & 1148   & 80.18 & 79.98 & 79.65 & 80.11 & 78.63 & 79.71 \\
        & SSFER-10\% & 2297   & 82.90 & 83.86 & 82.83 & 83.01 & 82.97 & 83.11 \\
        & SSFER-25\% & 5742   & 84.62 & \textbf{85.66} & \textbf{85.13} & \textbf{85.51} & 84.81 & 85.15 \\
    \bottomrule                         
    \end{tabular}
    }
\end{table}

\subsection{Comparison of param and FLOPs}
We directly adopted the results of the SOTA comparison in POSTER++ \cite{poster++}, and then added the baseline models for the comparison of parameters and FLOPs, the results are shown in Table \ref{tab flops}. 
We can see that the SSFER-25$\%$ outperforms the ViT-Large using 100$\%$ labels on all sides.
Although slightly weaker than the SOTA methods, the advantage of SSFER is mainly in data efficiency.
\begin{table}[H]
    \caption{\textbf{Comparison of Param and FLOPs on RAF-DB.}}
    \label{tab flops}
    \centering
    \resizebox{0.50\textwidth}{!}{
    \begin{tabular}{c c c c c }               
    \toprule                           
    \textbf{Method} & \textbf{Labels} & \textbf{Params (M)} & \textbf{FLOPs (G)} & \textbf{RAF-DB} \\
    \midrule                            
    DMUE        & 12271 & 78.4  & 13.4  & 89.42  \\
    TransFER    & 12271 & 65.2  & 15.3  & 90.91  \\
    POSTER-T    & 12271 & 52.2  & 13.6  & 91.36  \\
    POSTER-S    & 12271 & 62.0  & 14.7  & 91.54  \\
    POSTER      & 12271 & 71.8  & 15.7  & 92.05  \\
    POSTER++    & 12271 & 43.7  & 8.4   & 92.21  \\
    \midrule
    MobileNetV2 & 12271 & 2.2   & 0.3   & 83.08  \\
    MobileNetV3 & 12271 & 4.2   & 0.2   & 84.19  \\
    RestNet-18  & 12271 & 11.2  & 1.8   & 82.01  \\
    ResNet-50   & 12271 & 23.5  & 4.1   & 83.14  \\
    VGG11       & 12271 & 128.8 & 7.6   & 85.46  \\
    VGG13       & 12271 & 128.9 & 11.4  & 85.69  \\
    ViT-Tiny    & 12271 & 5.49  & 1.1   & 87.32  \\
    ViT-Small   & 12271 & 21.6  & 4.3   & 87.58  \\
    ViT-Base    & 12271 & 85.7  & 16.9  & 87.48  \\
    ViT-Large   & 12271 & 303.1 & 59.7  & 88.07  \\
    \midrule
    SSFER-5\%   & 614   & 85.7  & 16.9  & 80.96  \\
    SSFER-10\%  & 1227  & 85.7  & 16.9  & 85.07  \\
    SSFER-25\%  & 3068  & 85.7  & 16.9  & 88.23  \\
    \bottomrule                         
    \end{tabular}
    }
\end{table}

\subsection{Robustness and Generalization Analysis}
\textbf{Experiments on Adversarial Attack. }
To further validate the robustness and the importance of focused regions for emotion classification. 
We conducted an adversarial attack experiment on emotion focused and unfocused regions.

\begin{figure*}[htbp]
     \centering
     \includegraphics[width=6.2in]{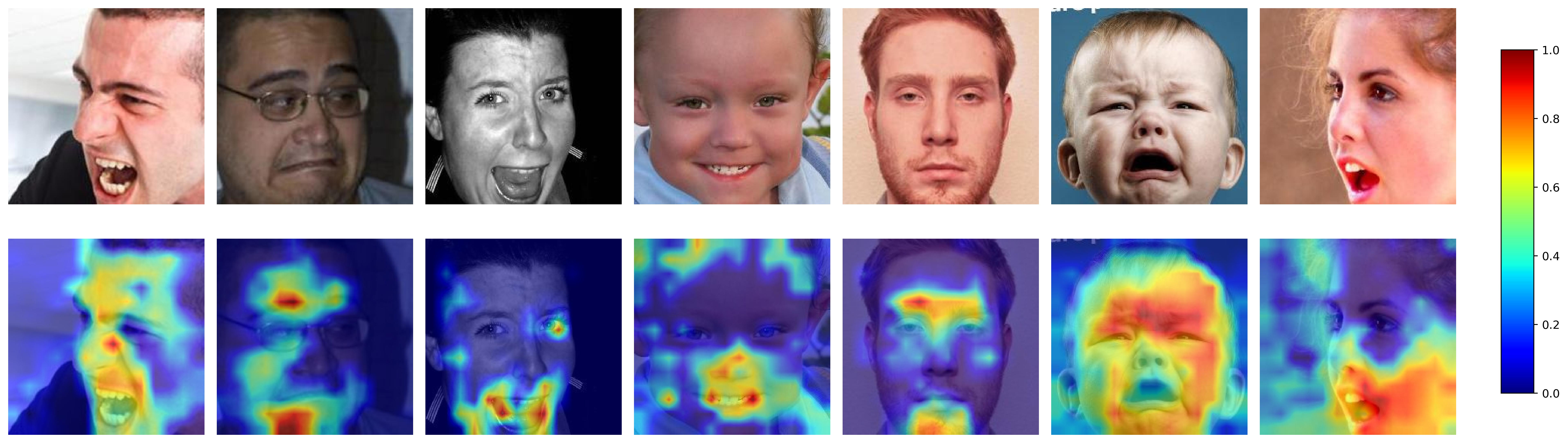}
     \caption{\textbf{The Class Activation Mapping (CAM) Obtained by Grad-CAM. } 
     The images are from the test set of RAF-DB. 
     Emotion focused regions are mainly concentrated in the eyes and mouth, which is highly consistent with the regions that humans focused on when recognizing facial expressions. }
     \label{fig focused regions}
 \end{figure*}

\begin{figure*}[htbp]
     \centering
     \includegraphics[width=6.2in]{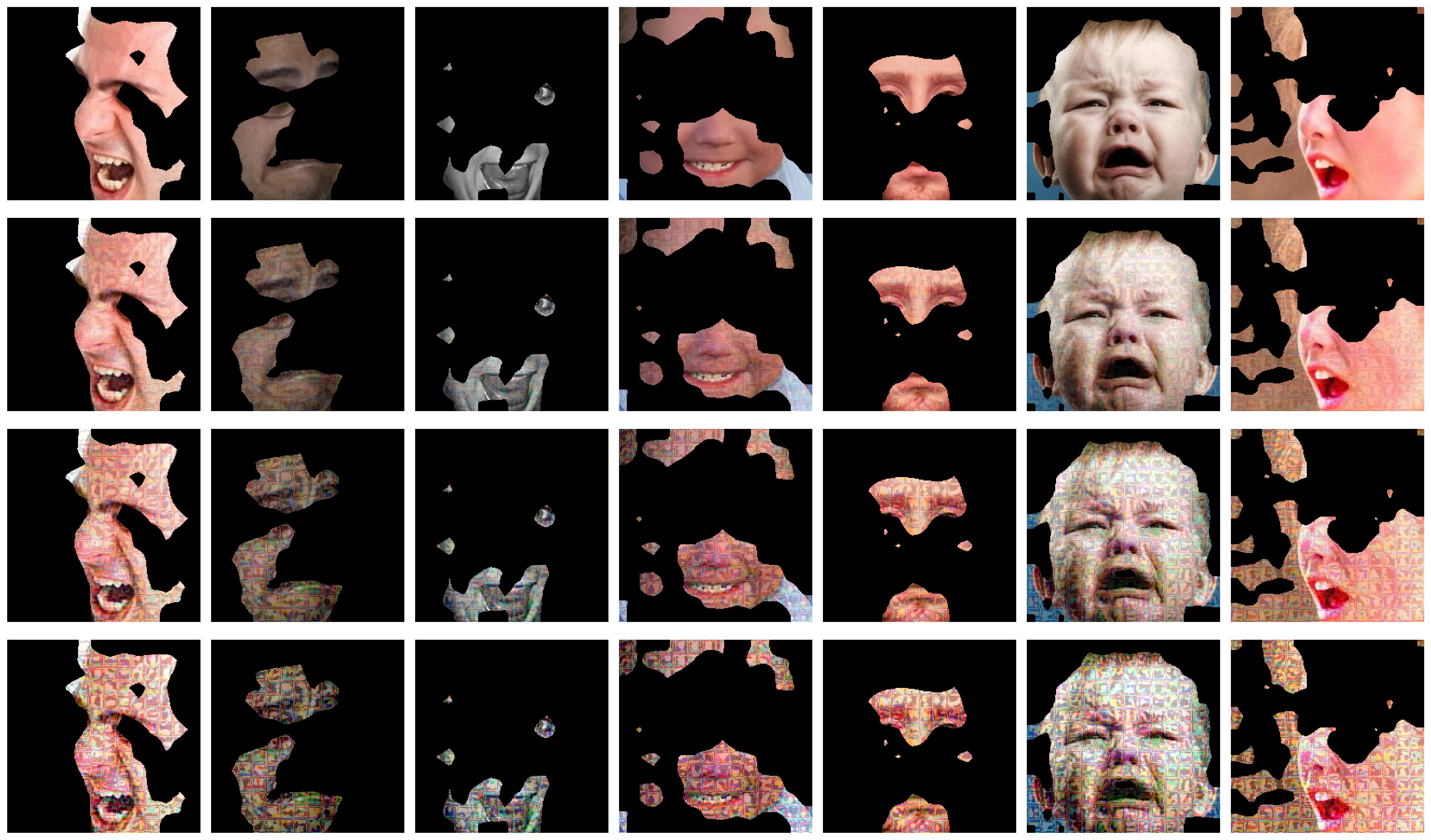}
     \caption{\textbf{Examples of FGSM Adversarial Attack on Emotion Focused Regions with Gradually Increasing Epsilons. } 
     Values above the threshold 0.3 are regarded as emotion focused regions, and other regions are regarded as emotion unfocused regions (black areas).}
     \label{fig fgsm attack}
 \end{figure*}

First, we used Grad-CAM \cite{Grad-CAM} to obtain the focused regions, as shown in Fig. \ref{fig focused regions}. 
Secondly, we conduct FGSM \cite{FGSM} attacks on both emotion focused and unfocused regions. 
The impact of the FGSM attack on the focused regions, with gradually increasing epsilon values, is shown in Fig. \ref{fig fgsm attack}.

The quantitative results in Table \ref{tab attack} show that there is a significant drop in accuracy when attacking the emotion focused regions obtained by Grad-CAM.
We observe that this decrease is significantly faster than attacks on emotion unfocused regions, validating the ability of SSFER to accurately identify emotion focused regions.

 \begin{table*}[htbp]
     \caption{\textbf{Results of FGSM Adversarial Attack on Emotion Focused and Unfocused Regions on RAF-DB, AffectNet, and FERPlus. } }
     \label{tab attack}
     \centering
     \resizebox{0.90\textwidth}{!}{
     \begin{tabular}{c c c c c c c c c}               
     \toprule                           
     \textbf{Attack regions} &\textbf{Method} & \textbf{Dataset} & \textbf{Eps=0} & \textbf{Eps=0.02} & \textbf{Eps=0.04} & \textbf{Eps=0.06} & \textbf{Eps=0.08} & \textbf{Eps=0.10} \\
     \midrule                            
     \multirow{4}{*}{emotion focused regions}&   SSFER-25$\%$    & RAF-DB        & 88.23 & 32.46 & 26.01 & 23.99 & 23.53 & 23.21   \\
     &                                           SSFER-25$\%$    & FERPlus       & 85.82 & 33.75 & 25.43 & 22.19 & 21.54 & 21.13   \\
     &                                           SSFER-10$\%$    & AffectNet-7   & 65.37 & 43.14 & 36.17 & 32.14 & 31.11 & 30.89   \\
     &                                           SSFER-10$\%$    & AffectNet-8   & 61.65 & 41.30 & 34.10 & 31.13 & 30.28 & 29.75   \\
 
     \midrule
     \multirow{4}{*}{emotion unfocused regions}& SSFER-25$\%$    & RAF-DB        & 88.23 & 47.03 & 39.77 & 37.22 & 36.08 & 35.85   \\
     &                                           SSFER-25$\%$    & FERPlus       & 85.82 & 46.54 & 38.24 & 35.13 & 34.26 & 33.98   \\
     &                                           SSFER-10$\%$    & AffectNet-7   & 65.37 & 49.77 & 42.29 & 39.89 & 38.91 & 38.57   \\
     &                                           SSFER-10$\%$    & AffectNet-8   & 61.65 & 47.23 & 40.08 & 38.45 & 37.88 & 37.23   \\
 
     \bottomrule                         
     \end{tabular}
     }
 \end{table*}

\textbf{Experiments on Label Noises. }
In Table \ref{tab noise labels}, we evaluate SSFER on RAF-DB and FERPlus at different levels of label noise to demonstrate its robustness. 
Here we directly used the results from Progressive Teacher \cite{jiang2021boosting} for comparison. 
The accuracy of SSFER decreases the least when the label noise on the RAF-DB ranges from 0$\%$ to 30$\%$.
\begin{table}[htbp]
    \caption{\textbf{Comparison of Different Label Noise Ratios on RAF-DB and FERPlus.}
    In SCN, "\ding{55}" Represent Fine-Tuning the Pre-Trained ResNet-18, "\ding{51}" Means SCN Algorithm is Used. }
    \label{tab noise labels}
    \centering
    \resizebox{0.55\textwidth}{!}{
    \begin{tabular}{c c c c c c c c}               
    \toprule                           
    \multirow{2}{*}{\textbf{Dataset}}    &\multirow{2}{*}{\textbf{Method}}  &\multirow{2}{*}{\textbf{Labels}}  &\multicolumn{4}{c}{\textbf{Label Noise Ratio}} &\multirow{2}{*}{\textbf{Decline $\downarrow$}}  \\
    \cline{4-7}
                        & & &\textbf{0}  &\textbf{10}  &\textbf{20}  &\textbf{30}\\
    \midrule                            
    \multirow{9}{*}{RAF-DB} 
        & SCN\ding{55}      &12271 &84.20  &80.81  &78.18  &75.26  &8.94   \\
        & SCN\ding{51}      &12271 &87.03  &82.18  &80.80  &77.46  &9.57   \\
        & RW Loss           &12271 &87.97  &82.43  &80.41  &76.77  &11.20  \\
        & SCAN-CCI          &12271 &89.02  &84.09  &78.72  &70.99  &18.03  \\
        & Mean Teacher      &12271 &88.41  &84.08  &81.40  &75.00  &13.41  \\
        & Progressive Teacher&12271&89.57  &87.28  &86.25  &84.32  &5.25   \\
        \cmidrule(lr){2-8}
        & SSFER-5\%         & 614   & 80.96  & 78.85  & 76.92  & 74.73  & 6.23  \\
        & SSFER-10\%        & 1227  & 85.07  & 83.12  & 81.26  & 79.47  & 5.60   \\
        & SSFER-25\%        & 3068  & 88.23  & 86.54  & 85.10  & 83.21  & \textbf{5.02}   \\
    \midrule
    \multirow{9}{*}{FERPlus} 
        & SCN\ding{55}      &12271  &86.80  &83.39  &82.24  &79.34  &7.46   \\
        & SCN\ding{51}      &12271  &88.01  &84.28  &83.17  &82.47  &5.54   \\
        & RW Loss           &12271  &87.60  &83.93  &83.55  &82.75  &4.85   \\
        & SCAN-CCI          &12271  &82.35  &79.25  &72.93  &68.90  &13.45  \\
        & Mean Teacher      &12271  &86.15  &82.87  &82.87  &72.06  &14.09  \\
        & Progressive Teacher&12271 &86.60  &85.07  &84.27  &83.73  &2.87   \\
        \cmidrule(lr){2-8}
        & SSFER-5\%         & 614   & 81.77  & 79.90  & 77.69  & 76.22  & 5.55   \\
        & SSFER-10\%        & 1227  & 83.95  & 82.60  & 81.11  & 79.81  & 4.14   \\
        & SSFER-25\%        & 3068  & 85.82  & 84.71  & 83.82  & 83.04  & \textbf{2.78}   \\
        \bottomrule                         
    \end{tabular}
    }
\end{table}

\textbf{Experiments on Other Tasks. }
The SSFER framework learns rich features of facial geometry and expression regions by reconstructing faces in the pre-training stage, which can be extended to other facial tasks. 
To validate the scalability of SSFER, we extended it to other facial tasks such as age classification and gender classification. 
K-fold cross-validation was conducted on the Audience \cite{Adience} dataset, and the results are shown in Table \ref{tab other tasks}.  
The results show that SSFER with 25$\%$ labels outperform ViT-Large with 100$\%$ labels on age and gender classification tasks, which is similar to the results on the FER task.  

\clearpage
\begin{table}[htbp]
    \caption{\textbf{Results of Gender Classification and Age Classification Tasks.}}
    \label{tab other tasks}
    \centering
    \resizebox{0.55\textwidth}{!}{
    \begin{tabular}{c c c c c c c c c}               
    \toprule                           
    \textbf{Task} &\textbf{Method} &\textbf{Labels} &\textbf{Fold1} &\textbf{Fold2} &\textbf{Fold3} &\textbf{Fold4} &\textbf{Fold5} &\textbf{Average}  \\

    \midrule                            
    \multirow{13}{*}{Gender} 
    &MobileNetV2    &8824    & 70.41 & 72.18 & 71.78 & 70.39 & 69.73 & 70.90 \\
    &MobileNetV3    &8824    & 70.86 & 71.93 & 72.99 & 69.98 & 70.19 & 71.19 \\
    &ResNet-18      &8824    & 71.42 & 73.70 & 73.65 & 70.54 & 71.10 & 72.08 \\
    &ResNet-50      &8824    & 70.30 & 72.99 & 73.50 & 74.20 & 71.66 & 72.53 \\
    &VGG11          &8824    & 72.89 & 74.52 & 75.43 & 73.44 & 73.48 & 73.95 \\
    &VGG13          &8824    & 73.55 & 74.26 & 74.21 & 73.18 & 73.59 & 73.76 \\
    &ViT-Tiny       &8824    & 77.21 & 77.11 & 79.70 & 77.96 & 77.45 & 77.89 \\
    &ViT-Small      &8824    & 80.00 & 78.53 & 80.25 & 78.21 & 78.01 & 79.00 \\
    &ViT-Base       &8824    & 80.05 & 80.51 & 82.18 & 79.99 & 80.14 & 80.57 \\
    &ViT-Large      &8824    & 80.71 & 81.62 & 81.31 & 80.65 & 81.56 & 81.17 \\
    \cmidrule(lr){2-9}
    & SSFER-5\%     &441     & 68.67 & 70.10 & 68.92 & 71.83 & 72.89 & 70.48 \\
    & SSFER-10\%    &882     & 77.74 & 78.90 & 76.11 & 76.15 & 78.38 & 77.46 \\
    & SSFER-25\%    &2206    & \textbf{82.25} & \textbf{82.15} & \textbf{81.49} & \textbf{82.40} & \textbf{81.73} & \textbf{82.00} \\
    \midrule
    \multirow{13}{*}{Age} 
    &MobileNetV2    &7878    & 56.75 & 53.94 & 53.90 & 55.85 & 55.89 & 55.27 \\
    &MobileNetV3    &7878    & 58.25 & 56.21 & 55.43 & 57.66 & 56.94 & 56.90 \\
    &ResNet-18      &7878    & 55.67 & 53.67 & 54.80 & 57.11 & 54.53 & 55.16 \\
    &ResNet-50      &7878    & 56.17 & 54.62 & 54.17 & 57.03 & 55.26 & 55.45 \\
    &VGG11          &7878    & 56.52 & 55.58 & 55.12 & 57.21 & 56.44 & 56.17 \\
    &VGG13          &7878    & 56.12 & 55.12 & 53.45 & 55.90 & 56.35 & 55.39 \\
    &ViT-Tiny       &7878    & 60.06 & 58.93 & 59.38 & 60.06 & 61.06 & 59.90 \\
    &ViT-Small      &7878    & 62.87 & 61.38 & 61.56 & 63.96 & 62.96 & 62.55 \\
    &ViT-Base       &7878    & 64.50 & 62.28 & 62.78 & 64.51 & 64.46 & 63.71 \\
    &ViT-Large      &7878    & 66.63 & 64.77 & 63.68 & \textbf{66.45} & \textbf{66.50} & 65.61 \\
    \cmidrule(lr){2-9}
    & SSFER-5\%     &394    & 54.59 & 52.69 & 51.39 & 52.09 & 52.69 &  52.81 \\
    & SSFER-10\%    &788    & 59.82 & 57.99 & 55.71 & 58.98 & 58.75 &  58.25 \\
    & SSFER-25\%    &1970   & \textbf{67.40} & \textbf{65.28} & \textbf{63.87} & 65.98 & 66.32 &  \textbf{65.77} \\
        \bottomrule                         
    \end{tabular}
    }
\end{table}

\subsection{Ablation Study}
\textbf{Analysis of proposed components.}
We stacked all components separately and step by step based on the pre-trained model to demonstrate their individual effectiveness. 
Based on the baseline, FaceMix is added in the supervised fine-tuning stage and EMA-teacher in the semi-supervised fine-tuning stage, respectively, and then they are combined, which constitutes the complete SSFER. 
We report the quantitative results on three datasets in Table \ref{tab ablation components}.
The results show that both FaceMix and EMA-Teacher are effective in improving the performance, validating the effectiveness of the proposed components.

\textbf{Analysis of self-supervised pre-training. }
To validate the effectiveness of self-supervised pre-training, we compare different pre-training paradigms and pre-training data as shown in Table \ref{tab pretrain compare}. 
The first two rows allow us to compare supervised pre-training with MAE \cite{MAE} pre-training, and the results show that MAE pre-training outperforms standard supervised pre-training, MAE is able to reconstruct the images and learn a more general feature representation. 
The last two rows allow us to compare the performance of MAE on different training data, and the results show that even though the facial features learned on the smaller Facial Images (1.2 million) are more effective than the general features learned on the large-scale ImageNet-21k (14 million).

\textbf{Analysis of semi-supervised learning framework.}
A comparison between Fixmatch and EMA-Teacher is presented in Table \ref{tab ablation semi combined}.
FixMatch shows much lower accuracy than EMA-Teacher, indicating that FixMatch is not a valid semi-supervised framework for ViT.

\textbf{Analysis of different mixing strategies. }
FaceMix and Mixup are compared and the results are shown in Table \ref{tab mixing strategies}. 
All the mixing strategies improve the results but FaceMix outperforms Mixup.
It is unsurprising, given that FaceMix is tailored for facial images.
By adjusting the loss weights of real and virtual images through IoU, it introduces diverse training samples while maintaining high-quality data for effective training.

\textbf{Analysis of $\kappa$ calculation. }
In the method section we mentioned that $\kappa$ was calculated not only using IOU, but also comparing the Peak Signal-to-Noise Ratio (PSNR), Structural Similarity Index (SSIM) \cite{SSIM}, and Feature Similarity Index (FSIM) \cite{FSIM}, and the ablation experiments for the $\kappa$ calculations are shown in Table \ref{tab ablation kappa}. 
The results show that IoU calculates $\kappa$ optimally, possibly because the other metrics (PSNR, SSIM, FSIM) are usually used to evaluate the overall image quality, focusing on overall information at the pixel level, and are not as effective as IoU is to boundaries and position.

\textbf{Analysis of loss functions. }
Four different FaceMix losses are defined. 
Each can be represented as follows: 
\begin{align}
    \mathcal{L}_1 & = (1-\kappa)\mathcal{L}_v + \mathcal{L}_i + \mathcal{L}_j \label{equ:loss1} \\
    \mathcal{L}_2 & = \mathcal{L}_v + (1-\kappa)(\mathcal{L}_i + \mathcal{L}_j) \label{equ:loss2} \\
    \mathcal{L}_3 & = \kappa\mathcal{L}_v + \mathcal{L}_i + \mathcal{L}_j \label{equ:loss3} \\
    \mathcal{L}_4 & = \mathcal{L}_v + \kappa(\mathcal{L}_i + \mathcal{L}_j) \label{equ:loss4}
\end{align}
We compare these four loss functions to demonstrate the effectiveness of our FaceMix loss, as shown in Table \ref{tab ablation loss}. 
The $\mathcal{L}_4$ performs the best, which is the FaceMix loss we finally adopted. 
This is because we want to ensure that our model is trained on high-quality samples while increasing diverse training samples.
When the differences in face angles and poses are small, we focus on training on virtual images and increase the number of high-quality training samples.
Conversely, when the differences are large, increase the ability to classify the original samples, and $\mathcal{L}_4$  follows this logic, resulting in better performance.

\textbf{Analysis of hyperparameter.}
Hyperparameter tuning plays a crucial role in optimizing model performance and we will analyze the effectiveness of hyperparameter tuning using the Grey Wolf Optimization (GWO) \cite{GWO}. 
The experimental results are presented in Table \ref{tab ablation GWO}. 
We optimized only a few parameters related to the learning rate (max\_learning\_rate, min\_learning\_rate, initial\_learning\_rate) with GWO in the supervised fine-tuning stage, and these hyperparameters were tuned to 1.23e-4, 3.29e-5 and 3.69e-5, respectively. 
The results show that tuning only the parameters related to the learning rate leads to improvements and that SSFER has great potential for optimization through hyperparameters. (We did not apply the new hyperparameters to other experiments. )

 \begin{table*}[htbp]
     \caption{\textbf{Ablation Study on Proposed Components on RAF-DB, AffectNet, and FERPlus.}
     The pre-trained model directly for standard supervised fine-tuning is set as the \textbf{baseline}.}        
     \label{tab ablation components}
     \centering
     \resizebox{0.85\textwidth}{!}
     {               
     \begin{tabular}{c cccc cccc cccc}               
     \toprule                           
     \multirow{2}{*}{\textbf{Method}} & \multicolumn{3}{c}{\textbf{RAF-DB}} & \multicolumn{3}{c}{\textbf{FERPlus}} & \multicolumn{3}{c}{\textbf{AffectNet-7}} & \multicolumn{3}{c}{\textbf{AffectNet-8}} \\                        
     \cline{2-13}    
         & \textbf{5$\%$} & \textbf{10$\%$} & \textbf{25$\%$} & \textbf{5$\%$} & \textbf{10$\%$} & \textbf{25$\%$} & \textbf{1$\%$} & \textbf{5$\%$} & \textbf{10$\%$} & \textbf{1$\%$} & \textbf{5$\%$} & \textbf{10$\%$} \\
     \midrule                            
         Baseline                 & 79.88 & 83.37 & 86.86 & 80.50 & 82.60 & 84.77 & 58.05 & 63.11 & 64.39 & 53.10 & 59.52 & 60.95 \\
         + FaceMix                & 80.57 & 84.41 & 87.84 & 81.29 & 83.27 & 85.37 & 58.85 & 63.91 & 65.17 & 53.62 & 59.90 & 61.25 \\
         + EMA-Teacher            & 80.18 & 84.25 & 87.51 & 81.06 & 83.03 & 85.12 & 58.42 & 63.45 & 64.77 & 53.25 & 59.77 & 61.40 \\
         + FaceMix + EMA-Teacher  & \textbf{80.96} & \textbf{85.07} & \textbf{88.23} & \textbf{81.77} & \textbf{83.95} & \textbf{85.82} & \textbf{59.08} & \textbf{64.02} & \textbf{65.37} & \textbf{53.85} & \textbf{60.17} & \textbf{61.65} \\
     \bottomrule                          
     \end{tabular}
     }
 \end{table*}

 \begin{table*}[htbp]                    
     \caption{\textbf{Comparison of ViT-Base in Different Pre-Training Paradigm and Datasets on RAF-DB, AffectNet, and FERPlus.}
     The supervised pre-training of the ViT-base was initialized using the standard ImageNet21K pre-trained and ImageNet1k fine-tuned weight provided by Steiner et al. \cite{steiner2021train}, which is publicly available in the timm PyTorch library.}       
     \label{tab pretrain compare}
     \centering
     \resizebox{0.90\textwidth}{!}
     {      
         \begin{tabular}{ccccccc}               
         \toprule                            
         \textbf{Model}&\textbf{Method}&\textbf{Pretrain Data}&\textbf{RAF-DB}&\textbf{FERPlus}&\textbf{AffectNet-7}&\textbf{AffectNet-8} \\
         \midrule                            
             ViT-B&SL    &ImageNet-21k    &86.47  &86.63  &60.14  &57.67    \\
             ViT-B&MAE   &ImageNet-21k    &88.49  &88.04  &63.05  &60.75    \\
             ViT-B&MAE   &Facial Images   &\textbf{91.19}  &\textbf{88.73}  &\textbf{66.62}  &\textbf{62.05}    \\
         \bottomrule                         
         \end{tabular}
     }
 \end{table*}

 \begin{table*}[htbp]
     \caption{\textbf{Ablation Study on Semi-Supervised Learning Framework on RAF-DB, AffectNet, and FERPlus.}}        
     \label{tab ablation semi combined}
     \centering
     \resizebox{0.90\textwidth}{!}
     {                 
     \begin{tabular}{c cccc cccc cccc}               
     \toprule                           
     \multirow{2}{*}{\textbf{Method}} & \multicolumn{3}{c}{\textbf{RAF-DB}} & \multicolumn{3}{c}{\textbf{FERPlus}} & \multicolumn{3}{c}{\textbf{AffectNet-7}} & \multicolumn{3}{c}{\textbf{AffectNet-8}} \\                        
     \cline{2-13}    
         & \textbf{5$\%$} & \textbf{10$\%$} & \textbf{25$\%$} & \textbf{5$\%$} & \textbf{10$\%$} & \textbf{25$\%$} & \textbf{1$\%$} & \textbf{5$\%$} & \textbf{10$\%$} & \textbf{1$\%$} & \textbf{5$\%$} & \textbf{10$\%$} \\
     \midrule                            
         FixMatch             & 70.17 & 75.26 & 80.73 & 77.05 & 79.63 & 81.44 & 42.94 & 49.48 & 50.57 & 34.27 & 40.52 & 43.62 \\
         EMA-Teacher          & \textbf{80.96} & \textbf{85.07} & \textbf{88.23} & \textbf{81.77} & \textbf{83.95} & \textbf{85.82} & \textbf{59.08} & \textbf{64.02} & \textbf{65.37} & \textbf{53.85} & \textbf{60.17} & \textbf{61.65} \\
     \bottomrule                         
     \end{tabular}
     }
\end{table*}

\begin{table*}[htbp]
     \caption{\textbf{Ablation Study on Mixing Strategies on RAF-DB, AffectNet, and FERPlus.}}        
     \label{tab mixing strategies}
     \centering
     \resizebox{0.90\textwidth}{!}
     {               
     \begin{tabular}{c cccc cccc cccc}               
     \toprule                           
     \multirow{2}{*}{\textbf{Method}} & \multicolumn{3}{c}{\textbf{RAF-DB}} & \multicolumn{3}{c}{\textbf{FERPlus}} & \multicolumn{3}{c}{\textbf{AffectNet-7}} & \multicolumn{3}{c}{\textbf{AffectNet-8}} \\                        
     \cline{2-13}    
         & \textbf{5$\%$} & \textbf{10$\%$} & \textbf{25$\%$} & \textbf{5$\%$} & \textbf{10$\%$} & \textbf{25$\%$} & \textbf{1$\%$} & \textbf{5$\%$} & \textbf{10$\%$} & \textbf{1$\%$} & \textbf{5$\%$} & \textbf{10$\%$} \\
     \midrule                            
         Mixup                  & 80.50 & 84.35 & 87.67 & 81.37 & 83.12 & 85.29 & 59.02 & 63.94 & 65.25 & 53.42 & 59.87 & 61.40 \\
         FaceMix                & \textbf{80.96} & \textbf{85.07} & \textbf{88.23} & \textbf{81.77} & \textbf{83.95} & \textbf{85.82} & \textbf{59.08} & \textbf{64.02} & \textbf{65.37} & \textbf{53.85} & \textbf{60.17} & \textbf{61.65} \\
     \bottomrule                          
     \end{tabular}
     }
 \end{table*}

 \begin{table*}[htbp]
     \caption{\textbf{Ablation Study on $\kappa$ Calculation on RAF-DB, AffectNet, and FERPlus.}}        
     \label{tab ablation kappa}
     \centering
     \resizebox{0.90\textwidth}{!}
     {               
     \begin{tabular}{c cccc cccc cccc}               
     \toprule                           
     \multirow{2}{*}{\textbf{Method}} & \multicolumn{3}{c}{\textbf{RAF-DB}} & \multicolumn{3}{c}{\textbf{FERPlus}} & \multicolumn{3}{c}{\textbf{AffectNet-7}} & \multicolumn{3}{c}{\textbf{AffectNet-8}}  \\                        
     \cline{2-13}    
         & \textbf{5$\%$} & \textbf{10$\%$} & \textbf{25$\%$} & \textbf{5$\%$} & \textbf{10$\%$} & \textbf{25$\%$} & \textbf{1$\%$} & \textbf{5$\%$} & \textbf{10$\%$} & \textbf{1$\%$} & \textbf{5$\%$} & \textbf{10$\%$} \\
     \midrule                            
         PSNR                &79.30  &83.14  &86.86  &80.88  &82.25  &83.92  &56.77  &62.23  &64.37  &51.35  &59.13  &59.23 \\
         SSIM                &79.46  &83.47  &86.57  &80.55  &82.43  &84.01  &57.03  &62.43  &64.26  &51.68  &58.95  &59.40 \\
         FSIM                &80.02  &83.64  &87.39  &81.62  &82.71  &84.27  &57.51  &63.06  &64.57  &52.05  &59.73  &60.75 \\
         IOU                 &\textbf{80.96}  &\textbf{85.07}  &\textbf{88.23}  &\textbf{81.77}  &\textbf{83.96}  &\textbf{85.82}  &\textbf{59.08}  &\textbf{64.03}  &\textbf{65.37}  &\textbf{53.85}  &\textbf{60.17}  &\textbf{61.65} \\
     \bottomrule                          
     \end{tabular}
     }
 \end{table*}

 \begin{table*}[htbp]
     \caption{\textbf{Ablation Study on Loss Functions on RAF-DB, AffectNet, and FERPlus.}}        
     \label{tab ablation loss}
     \centering
     \resizebox{0.90\textwidth}{!}
     {               
     \begin{tabular}{c cccc cccc cccc}               
     \toprule                           
     \multirow{2}{*}{\textbf{Method}} & \multicolumn{3}{c}{\textbf{RAF-DB}} & \multicolumn{3}{c}{\textbf{FERPlus}} & \multicolumn{3}{c}{\textbf{AffectNet-7}} & \multicolumn{3}{c}{\textbf{AffectNet-8}}  \\                        
     \cline{2-13}    
         & \textbf{5$\%$} & \textbf{10$\%$} & \textbf{25$\%$} & \textbf{5$\%$} & \textbf{10$\%$} & \textbf{25$\%$} & \textbf{1$\%$} & \textbf{5$\%$} & \textbf{10$\%$} & \textbf{1$\%$} & \textbf{5$\%$} & \textbf{10$\%$} \\
     \midrule                            
         $\mathcal{L}_1$            &79.79  &83.02  &86.28  &81.07  &82.47  &84.43  &56.06  &62.17  &63.09  &51.57  &58.57  &60.25 \\
         $\mathcal{L}_2$            &78.55  &82.89  &85.91  &80.53  &81.80  &83.22  &55.14  &61.37  &62.71  &50.35  &57.47  &59.53 \\
         $\mathcal{L}_3$            &78.62  &82.56  &85.85  &80.01  &81.81  &83.98  &55.37  &61.62  &62.31  &50.42  &57.02  &59.89 \\
         $\mathcal{L}_4$            &\textbf{80.96}  &\textbf{85.07}  &\textbf{88.23}  &\textbf{81.77}  &\textbf{83.95}  &\textbf{85.82}  &\textbf{59.08}  &\textbf{64.02}  &\textbf{65.37}  &\textbf{53.85}  &\textbf{60.17}  &\textbf{61.65} \\
     \bottomrule                          
     \end{tabular}
     }
 \end{table*}

 \begin{table*}[htbp]
     \caption{\textbf{Ablation Study on Hyperparameter Tuning Using GWO on RAF-DB, AffectNet, and FERPlus. }}        
     \label{tab ablation GWO}
     \centering
     \resizebox{0.90\textwidth}{!}
     { 
     \scriptsize
     \begin{tabular}{c cccc cccc cccc}               
     \toprule                           
     \multirow{2}{*}{\textbf{Method}} & \multicolumn{3}{c}{\textbf{RAF-DB}} & \multicolumn{3}{c}{\textbf{FERPlus}} & \multicolumn{3}{c}{\textbf{AffectNet-7}} & \multicolumn{3}{c}{\textbf{AffectNet-8}}  \\                        
     \cline{2-13}    
         & \textbf{5$\%$} & \textbf{10$\%$} & \textbf{25$\%$} & \textbf{5$\%$} & \textbf{10$\%$} & \textbf{25$\%$} & \textbf{1$\%$} & \textbf{5$\%$} & \textbf{10$\%$} & \textbf{1$\%$} & \textbf{5$\%$} & \textbf{10$\%$} \\
     \midrule                            
         SSFER                    &80.96  &85.07  &88.23  &81.77  &\textbf{83.95}  &\textbf{85.82}  &59.08  &64.02  &65.37  &53.85  &60.17  &61.65 \\
         SSFER(GWO fine-tuned)    &\textbf{81.29}  &\textbf{85.36}  &\textbf{88.89}  &\textbf{81.94}  &83.90  &85.81  &\textbf{59.82}  &\textbf{64.33}  &\textbf{65.48}  &\textbf{54.18}  &\textbf{60.60}  &\textbf{61.85} \\
     \bottomrule                          
     \end{tabular}
     }
 \end{table*}

\clearpage
\section{Discussion}
SSFER achieves significant improvements in semi-supervised FER by integrating self-supervised pre-training, semi-supervised fine-tuning, and an innovative data enhancement strategy, FaceMix. 
In this section, we discuss the reasons why SSFER is effective and its limitations, possible future directions, and conclude the research.

\textbf{The SSFER framework is effective mainly because: }
(i) SSFER makes better use of large-scale unlabeled FR datasets and learns rich features of facial geometry and expression regions through face reconstruction pre-training, providing a better initialization for subsequent two-stage fine-tuning.
(ii) FaceMix further alleviates the problem of the limited large and diverse datasets by adjusting loss weights of real and virtual images through IoU to add diverse training samples while ensuring that it is trained on high-quality samples. 
On the basis of these two elements, combined with the semi-supervised framework, SSFER achieves promising performance, and our extensive ablation experiments validate the effectiveness of each SSFER component.

\textbf{Despite the promising results, our study has some limitations. }
FaceMix improves the accuracy of the model but introduces some computational overhead due to the necessity of detecting face boxes.
Due to the limitation of computational resources, we only used ViT-Base for the experiments.
In addition, the class imbalance problem that exists in FER is not well coped by SSFER, resulting in a performance on the FERPlus dataset that is not as good as other datasets.

\textbf{Several promising directions can be explored in future study} based on our research results:
(i) Extension to other facial tasks, in our experiments, we have preliminarily verified that extending SSFER to age classification and gender classification can have encouraging results.
(ii) It is believed that expanding the pre-training data and upgrading the model to ViT-Large or even ViT-Huge will improve performance. 
(iii) Further expand the FER dataset by using an ensemble of well-trained FER models to label the FR dataset.

\section{Conclusion}
In this paper, we proposed a Self-supervised Semi-supervised Facial Expression Recognition (SSFER) framework to address the scarcity of large and diverse datasets in FER by exploiting large-scale FR datasets. 
The framework starts with face reconstruction pre-training on unlabeled facial images to learn general facial features and understand invariant patterns and relationships, such as facial geometry and expression regions. 
This is followed by supervised fine-tuning with FaceMix augmentation on labeled FER images, and finally, semi-supervised fine-tuning on unlabeled FER images. 
FaceMix plays a crucial role in our framework by generating diverse training samples and ensuring that our model is trained on high-quality images through the adjustment of loss weights based on their Intersection over Union (IoU).

SSFER effectively exploits the unlabeled FR dataset to improve FER performance, extensive experiments on RAF-DB, AffectNet, and FERPlus show that our method outperforms existing semi-supervised FER methods and achieves new state-of-the-art performance. 
Moreover, SSFER also demonstrates great scalability and robustness, establishing a strong benchmark for future research in semi-supervised FER, and providing a versatile and adaptable framework that can be extended to a variety of facial tasks, paving the way for further advancements in the field.

\section{Acknowledgments}
\noindent The research was supported by National Supercomputing Center in Chengdu.

\bibliographystyle{elsarticle-num}
\bibliography{neurocomputing_manuscript}
\end{document}